\documentclass[pdflatex,sn-mathphys-num]{sn-jnl}% Math and Physical Sciences Numbered Reference Style 
\usepackage{graphicx}%
\usepackage{multirow}%
\usepackage{amsmath,amssymb,amsfonts}%
\usepackage{amsthm}%
\usepackage{mathrsfs}%
\usepackage[title]{appendix}%
\usepackage{xcolor}%
\usepackage{textcomp}%
\usepackage{manyfoot}%
\usepackage{booktabs}%
\usepackage{algorithm}%
\usepackage{algorithmicx}%
\usepackage{algpseudocode}%
\usepackage{listings}%
\usepackage{subfigure}
\theoremstyle{thmstyleone}%
\theoremstyle{thmstyletwo}%

\theoremstyle{thmstylethree}%

\begin{document}

\title[Article Title]{Accurate Multi-Category Student Performance Forecasting at Early Stages of Online Education Using Neural Networks}


\author[1,2]{\fnm{Naveed Ur Rehman} \sur{Junejo}}\email{naveed.rehman@hstc.edu.cn}

\author[3]{\fnm{Muhammad Wasim} \sur{Nawaz}}\email{muhammad.wasim@dce.uol.edu.pk}

\author*[4]{\fnm{Qingsheng} \sur{Huang}}\email{huangqs@hstc.edu.cn}

\author[1]{\fnm{Xiaoqing} \sur{Dong}}\email{dxqzq@hstc.edu.cn}

\author[1]{\fnm{Chang} \sur{Wang}}\email{wchang@hstc.edu.cn}

\author*[2]{\fnm{Gengzhong} \sur{Zheng}}\email{zhenggz@hstc.edu.cn}

\affil[1]{\orgdiv{School of Physics and Electronic Engineering}, \orgname{Hanshan Normal University}, \orgaddress{\street{} \city{Chaozhou}, \postcode{521041}, \state{Guangdong}, \country{China}}}

\affil[2]{\orgdiv{Department of Computer Science and Engineering}, \orgname{Hanshan Normal University}, \orgaddress{\street{} \city{Chaozhou}, \postcode{521041}, \state{Guangdong}, \country{China}}}

\affil[3]{\orgdiv{Department of Computer Engineering}, \orgname{The University of Lahore}, \orgaddress{\street{} \city{Lahore}, \postcode{54000}, \state{Punjab}, \country{Pakistan}}}

\affil[4]{\orgdiv{School of Mathematics and Statistics}, \orgname{Hanshan Normal University}, \orgaddress{\street{} \city{Chaozhou}, \postcode{521041}, \state{Guangdong}, \country{China}}}


\abstract{The ability to accurately predict and analyze student performance in online education, both at the outset and throughout the semester, is vital. Most of the published studies focus on binary classification (Fail or Pass) but there is still a significant research gap in predicting students’ performance across multiple categories. This study introduces a novel neural network-based approach capable of accurately predicting student performance and identifying vulnerable students at early stages of the online courses. The Open University Learning Analytics (OULA) dataset is employed to develop and test the proposed model, which predicts outcomes in Distinction, Fail, Pass, and Withdrawn categories. The OULA dataset is preprocessed to extract features from demographic data, assessment data, and clickstream interactions within a Virtual Learning Environment (VLE). Comparative simulations indicate that the proposed model significantly outperforms existing baseline models including Artificial Neural Network Long Short Term Memory (ANN-LSTM), Random Forest (RF) ‘gini’, RF ‘entropy’ and Deep Feed Forward Neural Network (DFFNN) in terms of accuracy, precision, recall, and F1-score. The results indicate that the prediction accuracy of the proposed method is about $25\%$ more than the existing state-of-the-art. Furthermore, compared to existing methodologies, the model demonstrates superior predictive capability across temporal course progression, achieving superior accuracy even at the initial $20\%$ phase of course completion.

}




\maketitle

\section{Introduction}\label{sec1}
Predicting and understanding student achievement is one of the biggest challenges faced by educators. Online learning environments continue to struggle with the issue of high dropout rates, making accurate predictions crucial for reducing this rate \citep{a1, a2, a3, b10, b13}. The primary goal of the current work is to develop a predictive model to identify students who are at risk of dropping out early. This prediction model will be extremely helpful to instructors, enabling them to make timely and effective interventions, which will reduce the dropout rate. Massive Open Online Courses (MOOCs) and other online education platforms have grown in popularity recently, offering students more opportunities to acquire learning materials and advance their careers. However, a significant concern is the high dropout rate associated with these online courses \cite{a1, a6, a7}. MOOCs have become increasingly important in education because they provide students with access to resources that help them learn and develop their skills.

Even with the advantages of virtual learning environments, high dropout rates remain a problem. It makes sense that, in order to lower dropout rates and increase course continuity, educators and academics are highly interested in developing models and tactics that can assess students' behavior and personal data. Identifying at-risk students who may be likely to drop out can be challenging using traditional educational methodologies, making it difficult to offer effective interventions \cite{a8, a9}. Various models and tactics have been proposed by researchers to anticipate students who are at risk of dropping out in an attempt to address this issue \cite{b10, b13, b4, b5, b6, b12, b18, b20}. However, these existing approaches can be tedious and time-consuming and often depend on manually created features. To address these issues, recent research has suggested building predictive models for student performance and dropout prediction using machine learning and deep learning algorithms.

Achieving analytics objectives is made easier by understanding the performance of a class of students during the first few days of a course in virtual learning environments (VLE). Numerous academics have suggested predictive models to forecast student success in Massive Open Online Courses (MOOC) in binary classifications (pass or fail) or (dropout or not) \cite{b4, b5, b1, b2, b3, b7, b9, b8}. Nevertheless, a limited number of research have created models that forecast student performance in multi-classification formats, such as \cite{b10, b20, b15, b17, a10, a11}. Therefore, further research is required to enhance the prediction performance of multi-class models \cite{b20, a12}. While more sophisticated techniques like deep learning are rarely used, traditional artificial intelligence techniques are frequently employed in predicting student success in online higher education \cite{a13}. Since most research has focused on creating prediction models for specific courses, such as \cite{b12, b11, a15}, overfitting may occur if new courses are created \cite{a16}. Therefore, it is necessary to develop predictive models that can learn and evaluate various courses to address this issue..

The lack of study on multiclass classification with the OULAD dataset is due to difficulties in attaining high accuracy because performance is frequently deteriorated by the additional difficulty of differentiating between many classes. The complex correlations between various variables from different dataset files, which are essential for precise multiclass predictions, are frequently not adequately captured by current approaches. Closing this gap is critical because multiclass categorization can offer a more thorough and practical picture of student performance, enabling individualized support and focused interventions. By developing a novel data preprocessing pipeline and cutting-edge feature engineering techniques, our work closes this gap by improving the integration and distillation of information from multiple datasets and boosting the predictive power of models for multiclass classification tasks within the OULAD framework.
Main contributions of this work include: 
\begin{itemize}
\item We propose a novel data preprocessing pipeline that effectively merges and aggregates multiple dataset files, retaining crucial features.
\item Our method integrates diverse dataset files, providing a unified view of student data and enabling more accurate predictions through the inclusion of previously unexplored feature relationships.
\item Our approach introduces innovative feature engineering techniques that distil data from multiple dataset files into informative attributes, enhancing the accuracy of early student performance prediction.
\item We present and utilize a one dimensional Convolutional Neural Network (1D-CNN) model for multiple categories that achieves outstanding accuracy, precision, recall, and F1-score in predicting student performance in online learning environments. Furthermore, we also analyse our method in term of Area Under the Curve (AUC), where the high AUC values reflect our proposed method's excellent ability to distinguish between different levels of students achievement.
\item  By accurately early predicting student performance at about $20\%$ of the course length, our approach facilitates timely interventions to improve learning outcomes in online education environments, our proposed model attains $92\%$ of accuracy. 
\item Experimental results show that the proposed model outperforms compared models including ANN-LSTM, RF 'gini, RF 'entropy', and DFFNN in terms of precision and recall of $98\%$.
\end{itemize}
This paper structure is organized as follows. In Section 2 the detailed literature review of the students performance prediction is presented. The explanation of the proposed methodology which contains the detailed description of dataset, data preprocessing, and proposed 1D CNN model is mentioned in Section 3. In Section 4, the results and discussion of our study are analyzed. In Section. 5, conclusion, an overview of the findings, and suggestions
for future research is provided.
\iffalse
The Introduction section, of referenced text \cite{bib1} expands on the background of the work (some overlap with the Abstract is acceptable). The introduction should not include subheadings.

Springer Nature does not impose a strict layout as standard however authors are advised to check the individual requirements for the journal they are planning to submit to as there may be journal-level preferences. When preparing your text please also be aware that some stylistic choices are not supported in full text XML (publication version), including coloured font. These will not be replicated in the typeset article if it is accepted. 
\fi

\section{Literature Work of Student Performance Prediction}
Wasif et al. provided a way to use logging data from the OULAD to predict student performance in their 2019 study. For comparison in binary classification, they used Gaussian Naïve Bayes, random forest (RF), logistic regression (LR), and multilayer perceptron (MLP) models. The findings showed that the RF model performed better than the others, with $89\%$, $89\%$, $88\%$, and $88\%$ for accuracy, precision, recall, and F1-score, respectively. The models' ability to accurately identify children who are at-risk allows for prompt interventions that improve student progress, according to the authors' conclusion \cite{b1}.
In 2019, Hassan et al. proposed a deep learning model named the Long Short-Term Memory (LSTM) deep model to predict withdrawn students using only clickstream data of OULAD. A binary classification problem has been considered where the “pass” class has been merged with the “distinction” to formulate one class and the other was the “withdrawn” class. 
The accuracy was approximately $80\%$ in the fifth week and approximately $97\%$ in the twenty-fifth. The proposed model namely LSTM outperforms LR and artificial neural networks in terms of accuracy, precision, and recall \cite{b2}.
In 2019, Aljohani et al. used a deep learning model to identify students who were likely to fail a course, allowing instructors to provide the appropriate support on the clickstream data. 
The suggested model has a $90\%$ accuracy rate in predicting pass and fail grades in the first 10 weeks of student engagement in a virtual learning environment (VLE). In the pass and fail classification, the LSTM model superseded the baseline LR and ANN in terms of precision and recall value by $93.46\%$ and $75.79\%$, respectively \cite{b3}. 

In 2019, Waheed et al. proposed the deep learning ANN model to predict the students’ at-risk whether they are distinct, withdrawn, and fail by utilizing the clickstream and demographic data. The problem has been treated as a binary classification for the final result. Out of a total of 54 features, the authors employed sparse reduction to choose 30 features, and then they transformed the values of the selected features using the "MinMax" scaler. In the fail prediction model, withdrawn student’s data was ignored, and distinction and pass data were merged, where the accuracy obtained in the fail prediction model in quarter1, quarter2, quarter3, and quarter4 was around $77\%$, $81\%$, $86\%$, and $88\%$, respectively. Fail data was overlooked, and distinction and pass data have been considered as one class in the scenario of withdrawn prediction. In quarter 1, quarter 2, quarter 3, and quarter 4, the accuracy attained in the withdrawn prediction model was about $78\%$, $86\%$, $90\%$, and $93\%$, respectively. ANN model results attained a better performance in comparison of the baseline LR and support vector machine (SVM) models in terms of accuracy \cite{b4}. 
In 2020, He et al. proposed the joint deep neural network namely recurrent neural network (RNN) - gated recurrent unit (GRU) to fit both static and sequential data, to fill the missing stream data, the data completion mechanism has also been implemented. 
For online course outcome prediction, the author did a binary classification. The classes ‘pass’ and ‘withdrawal’ were ignored; ‘pass’ and ‘distinction’ were considered. Results revealed that methods such as RNN and GRU outperformed LSTM. 
As a result, the suggested RNN-GRU achieved over $80\%$ accuracy in predicting the students who will be at risk at the conclusion of the semester. The forecast applied to the entire course. Thus, the most recent courses served as a test set. Throughout all courses, the average accuracy from the fifth to the thirty-ninth week ranged from $60\%$ to $90\%$ \cite{b5}. 

Chui et al. (2020) developed a Reduced Training Vector-Based Support Vector Machine (RTV-SVM) classifier to forecast marginal or at-risk pupils based on academic performance, taking into account the classifier's longer training period. 
The data of 32,593 students in the OULAD includes session logs documenting students' interactions with the VLE system and student demographic data were used. While drastically cutting the training time by $59\%$, TRV-SVM predicted the student at risk and marginal student with good accuracy, scoring $93.8\%$ and $93.5\%$, respectively \cite{b6}.
In order to anticipate students' final exam marks in 2021, Kumar et al. assessed the effectiveness of classification models such as RF, NB, SVM, k-NN, and ANN utilizing clickstream, assessment, and demographic data as input of the prediction algorithms \cite{b7}. Esteban (2021) investigated the significance of assignment data for predicting students' performance. A multiple-instance learning prediction algorithm was suggested by the authors to predict which students will pass and fail. The model's input has been assessment data \cite{b8}. In 2021, Hlioui et al. used assessment, clickstream, and demographic predict withdrawal rates in students using a variety of models, including Random Forest (RF), Decision tree (J48), MLP, SVM, and Bayesian (TAN) classifiers. Two classes of performance prediction have been identified: withdrawal and completion (sometimes known as "Distinction", "Pass", or "Fail") \cite{b9}.

In 2021, Adnan et al. employed a Deep Feed Forward Neural Network (DFFNN) to estimate the final grading of students such as pass, distinct, fail, and withdrawn. The greatest average accuracy recorded at the end of the course was $43\%$, $63\%$, $71\%$, and $72\%$ when the input data consisted of demographic information, demographic information and click steam, demographic information, click stream, and assessment, and all features, respectively. 
By the end of the courses, $90\%$ accuracy had been achieved for the final grades while considering the binary classification \cite{b10}. 
A semi-supervised learning ensemble model of an artificial neural network (ANN) was proposed by Vo et al. in 2021 to forecast student performance (pass or fail) before to the midterm and at the end of the course. 
Five different courses including AAA, BBB, CCC, DDD, and FFF have been considered, and these courses were selected to be inputs into the proposed model. Every selected subject was offered in two distinct semesters. While the second course was used for testing, the first one was used for training. Averaging $87.47\%$, the chosen courses' accuracy during the midterms of the semester. About $40\%$ of the OULAD student data were disregarded when it came to distinction and withdrawal. The proposed ANN model has been compared to baselines KNN, RF, NB, C4.5, Logistic, and SVM \cite{b11}. 
To anticipate whether a student would stay or drop out the previous week, in 2021 Pei et al created supervised machine learning models such as SVM, RF, and DT using the expectation maximum technique. Then, it used the resultant probability as input to increase the prediction accuracy for the current week. It used aggregated clickstream data and demographic data of two STEM courses (CCC-2014J and CCC-2014B) and two social science courses (AAA-2014J and AAA-2013J) as input. A Synthetic Minority Over-Sampling (SMOTE) technique was used for balancing the data. For the chosen courses, the average accuracy throughout all weeks was almost $88\%$ \cite{b12}.

In 2021, Waheed et al. implemented a deep learning-based approach named sequential conditional generative adversarial network (SC-GAN) generates synthetic student records for the next timestamp by encapsulating each student's past behavior for its previous sequences. The authors included the "fail" and the updated "pass" classes in their binary classification to predict students' learning performance. 
The experimental results validated that the SC-GAN outperformed Synthetic Minority Oversampling and conventional Random Over-sampling in terms of area under the curve (AUC) of $6.53\%$ and $7.07\%$, respectively \cite{b13}. In 2022, Qiu et al. presented a prediction framework based on six classic machine learning algorithms including SVC (R), SVC (L), Naïve Bayes (NB), K-Nearest Neighbor (KNN), and sofmax. The predictor's projected output value was either qualified or unqualified. In 12 activities, the authors employed click stream data for “DDD” course data \cite{b14}.

In 2022, Adnan et al. considered a multiclass classification scenario with four classes including pass, distinct, withdraw, and fail to predict student performance. Authors implemented deep learning and ML models such as FFNN, RF with two different criteria ‘gini’ and ‘entropy’, SVM, DT, KNN, LR, AdaBoost, Gradient Boosting, Bernoulli Naïve Bayes, Gaussian Naïve Bayes, and Extra Tree classifier to compare their performance. 
Experimental results witnessed that the RF ‘gini’ and RF ‘entropy’ outperformed in terms of accuracy as compared other models \cite{b15}. 
When employing clickstream and demographic data, multi-class student performance prediction models still require to improve their accuracy early in MOOC courses. Yanqing Xie in 2021 a proposed model named; Attention-based Multi-layer LSTM (AML) to predict student performance, which combines student demographic data and clickstream data as input of the model for comprehensive analysis \cite{b16}. Although, the predictive models proposed by Adnan (2021) \cite{b10} and (2022) \cite{b15} are multiclass classification models, these methods were not used early in the course period, the accuracy attained at the end of the course were $63\%$ and $66\%$, respectively, which can be considered low results \cite{b16}. Yanqing Xie has considered two scenarios; one is four-class classification and the other is binary classification. The accuracy obtained for four-class classifications from week 0, week 5, and week 25 were $43.93\%$, $53.51\%$, and $57.40\%$, respectively. The accuracy attained for binary classification from week 0, week 5, and week 25 were $65.62\%$, $78.94\%$, and $95.75\%$, respectively. Experimental results have proven that the AML model attained better accuracy then state-of-the-art models for binary as well as four-class classifications \cite{b16}. 

As a multiclass classification model, Hao Jia et al. in 2022 proposed a students’ performance prediction network (SPBN) model to predict the performance of the students in the final, only assessment data has been considered as input. In addition, researchers compared it with different classical machine learning models. Based on the students' assessment results, the computed performance value was split into four categories: fail, pass, good, and distinction. 
The average F1-score achieved by SPBN is about $86.6\%$, while the same metrics are about $81.4\%$, $80.90\%$, $74.9\%$, $85.5\%$, $71.3\%$, and $55.2\%$ for LightGBM, XGBoost, AdaBoost, RF, MLP and Naïve Bayes, respectively \cite{b17}.
In 2023, Waheed et al. presented the results of a study that examined how well-known deep learning technology LSTM performed in predicting which students would fail a course that was given in an online, self-paced learning environment. 
Surprisingly, the LSTM algorithm achieves up to $71\%$ accuracy with only the first five weeks of course activity log data used for training to distinguish between pass and fail classes. This surpasses nearly all other traditional algorithms, even though the LSTM algorithm is trained on the entire dataset collected for the course, i.e., up to 38 weeks \cite{b18}.

In 2023, Nafea proposed an approach using group modeling to identify the students’ performance. Authors have merged the strengths of machine learning algorithms namely DT, SVM, AdaBoots, and RF. %Then, because it takes into account, the final ensemble estimates were used as one of the best LR techniques to produce a strong and trustworthy predictive model. 
In this study, the "pass" and "distinct" labels were merged and allotted as the "pass" label. Similarly, the "fail" and "withdrawn" labels were merged, and allocated the "fail" label. %A total of 15,385 pupils were classified as "pass" and 17,208 students as "fail". 
The results have proven that the proposed idea outperforms the baseline by attaining $95\%$ accuracy \cite{b19}. 

Fatima Ahmed Al-Azazi in 2023 proposed a deep learning-based artificial neural network combined with Long Short-Term Memory, namely ANN-LSTM multiclass model to predict students’ performance \cite{b20}. The authors compared their proposed ANN-LSTM model with two baseline methods including the Gated Recurrent Unit (GRU) and Recurrent Neural Network (RNN). The accuracy attained at the end of the third month of the semester by ANN-LSTM, GRU, and RNN were about $70\%$, $57\%$, and $53\%$, respectively. Results witnessed that ANN-LSTM outperformed the baseline models in terms of accuracy. Furthermore, authors also compared ANN-LSTM with DFFNN proposed by Adnan in 2021 \cite{b10}, the ML model named; RF ‘gini’ and RF ‘entropy’ proposed by Adnan in 2022 \cite{b15}, and the AML model proposed by Xie in 2021 \cite{b16} have been considered as baseline models because they have executed multiclass classification which are closely related to the authors’ research work. In the testing data, on the first day and 270th day (last day) of the course the accuracy was achieved almost $43\%$ and $72\%$, respectively \cite{b20}. 

\section{Methodology}
\label{sec:sample3}
In this section, we present the VLE dataset details, its pre-processing, and building and training of the neural network model. Figure 1 depicts the block diagram of our experimental workflow. 

\begin{figure}[h!]
	\centering
	% \resizebox{.49\textwidth}{.17\textheight}{%
    % \scalebox{1.2}{\includegraphics[width=0.85\linewidth]{Block Diagram of VLE1.pn}
		\includegraphics[width=0.87\linewidth]{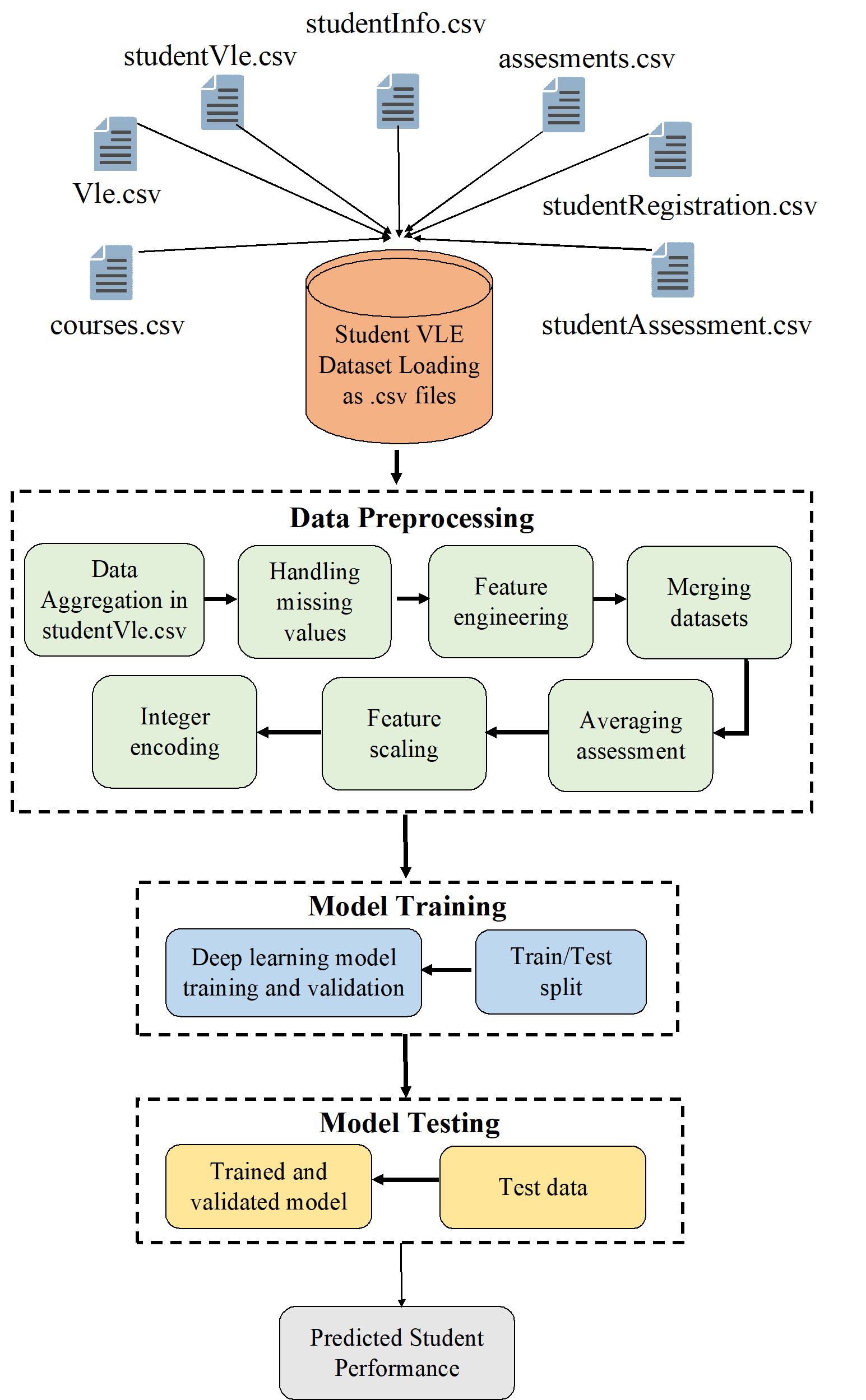}
	\caption{Block diagram of the proposed framework.}
        \label{fig:fig1 block diagram}
\end{figure}

\subsection{Dataset}
For training machine learning models, we employed the Open University Learning Analytics Dataset (OULAD), which contains details about the students' demographics, course registration, interactions on the VLE, clickstreams, assessments, and final performance scores \cite{OULAD2017data}. The OULAD is free, publicly available, and officially verified by the Open Data Institute (ODI): http://theodi.org/. In the obtained OULAD, the clickstream, course information, student details, and registration data are organized in seven tables, with the students serving as the main informational focus. This dataset contains data from 32,593 students collected over nine months. Four classes/grades are created from the students' final performance including Fail, Distinction, Pass, and Withdrawn.
Students had access to seven courses or modules, which were offered at least twice a year at different times. Seven tables comprising the obtained raw data are linked together by identifying columns. The tables include data on assessments, types of assessments, due dates for submissions, courses, types of VLE materials, clickstreams showing how students engage with the VLE, course registration details, and student demographics. The VLE clickstream contains details on the different activities that students do on heterogeneous modules, like using discussion boards, getting course materials, logging in, and visiting the main page, subpage, URL, and glossary of OU, among other things. The students' level of interest in each sort of activity is recorded by their clickstream activity.

\subsection{Data Pre-processing}
This section outlines the steps taken to pre-process OULAD and perform feature engineering to early predict student performance. After loading the dataset, we aggregate and merge various dataset files. The Virtual Learning Environment (VLE) data file contains information about students' interactions with the online learning platform, such as the number of clicks on course materials. These interactions are indicative of students' engagement levels.
\subsubsection{Data aggregation} It helps in summarizing information and reducing the dataset's dimensionality while retaining important insights. We aggregate the VLE data by grouping it based on the $code\_module$, $code\_presentation$, $ID\_student$, and $date$, then summing the total number of clicks for each group. This aggregation results in a new feature, $total\_clicks$, which represents the engagement of each student on each day.
\subsubsection{Handling missing values} It ensures that the dataset is clean and enriched with relevant information for the model. Missing values can occur due to various reasons, such as incomplete data collection or errors during data entry. In our dataset, we handle missing values of $unregistration\_date$ by filling them with a specific value (in our case, 270). This approach helps maintain the dataset's integrity without dropping potentially valuable rows.
\subsubsection{Feature engineering} It involves creating new features from existing data to improve the predictive power of the model. One such feature is the total number of days a student was registered for a course namely $total\_reg\_days$. This feature is derived by subtracting the $registration\_date$ from the $unregistration\_date$ in $student\_registration.csv$.
\subsubsection{Merging datasets} It is essential to combine various sources of information into a single, cohesive dataset, facilitating comprehensive analysis. We merge the $student\_info.csv$, $student\_registration.csv$, and $student\_vle.csv$ dataset files based on common columns ($code\_module$, $code\_presentation$, and $id\_student$). This merging process integrates demographic information, registration details, and engagement metrics into a single dataset. This comprehensive dataset allows us to have all relevant features for each student in one place, making it easier to analyze and model their performance. Assessment weights indicate the importance of each assessment in the overall grading scheme of a course. Calculating the average weights helps in understanding the assessment structure across different modules and presentations. We group the $assessments.csv$ dataset file by $code\_module$ and $code\_presentation$, and then calculate the mean weight of the assessments within each group. This aggregation provides an average weight for assessments in each module and presentation. The calculated average weights are merged with the $student\_info.csv$ dataset file to include this information as a new feature namely weight.
\subsubsection{Encoding} We convert the categorical features into numerical values. We also perform integer label encoding to assign a unique integer to each category. 0: Distinction,  1: Fail, 2: Pass, 3: Withdrawn.
\subsubsection{Feature scaling} It standardizes the range of independent variables or features by bringing all features to a similar scale, improving the performance and training stability of the model. We use the StandardScaler to standardize features by removing the mean and scaling to unit variance. 
\subsubsection{Correlation analysis} we calculate the Pearson Correlation Coefficient, which is a measure of the linear correlation between two variables. It ranges from -1 to 1, where 1 indicates a perfect positive linear relationship, -1 indicates a perfect negative linear relationship, and 0 indicates no linear relationship. We extract the correlation values between each feature and the target variable final\_result. 
\subsubsection{Dataset filtering for early performance prediction} To facilitate early prediction of student performance, we filter the dataset to include only the initial period of the course using the sorted date column to ensure chronological order. We then calculate the index corresponding to the specified percentage of the course duration, e.g., $5\%$, $10\%$, $20\%$, $40\%$, $60\%$, $80\%$, and $100\%$.
\subsubsection{Dataset splitting} We split the dataset into training ($70\%$) and testing ($30\%$) sets. We also ensure that the split is stratified based on the target variable to maintain the distribution of classes.
\subsubsection{Class weights} These are defined to handle class imbalance, ensuring that the model does not bias towards the majority class. While the weights can be chosen based on the inverse frequency of each class, we found the following simple class weighting scheme useful in our case: 
Distinction = 1.5,  Fail = 1.5,  Pass = 1, and Withdrawn=1.  
We provide these weights to the model during training to emphasize underrepresented classes.
\subsection{Building and training the 1D-CNN model} 
\begin{figure}[h!]
	\centering
	% \resizebox{.49\textwidth}{.17\textheight}{%
		\includegraphics[width=1.0\linewidth]{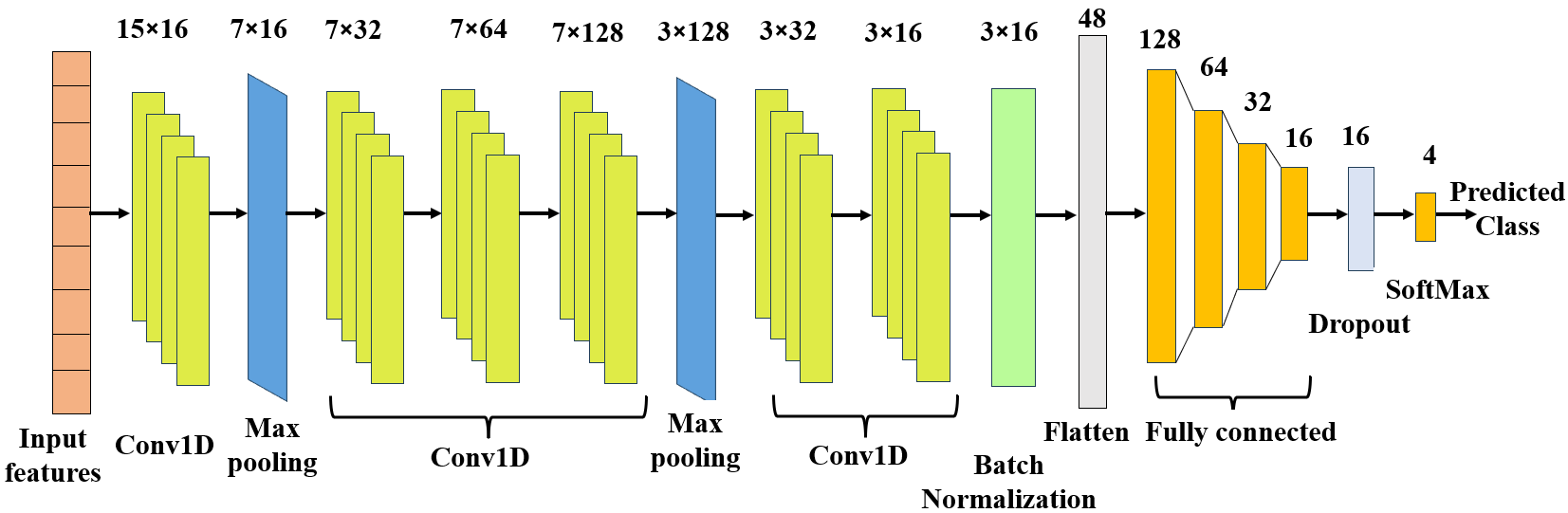}%}
	\caption{1D Convolutional Neural Network Model.}
        \label{fig:fig2 1D Convolutional Neural Network Model}
\end{figure}

We build and train a one-dimensional Convolutional Neural Network (1D-CNN) model for predicting student performance as depicted in Figure \ref{fig:fig2 1D Convolutional Neural Network Model}. The model architecture includes convolutional layers for feature extraction, fully connected layers for classification, and various techniques like batch normalization and dropout for improving performance and preventing overfitting.

\begin{itemize}
\item Compiling the model: After defining the architecture, we compile the model using the sparse categorical cross-entropy loss, Adam optimizer, and accuracy metric. 
\item Training the model: We train the model on the training data, specifying the number of epochs = 70, batch size = 1024, and validation split = 0.1.  
\end{itemize}
The model is trained using the processed data, and its performance is evaluated based on accuracy, precision, recall, and F1 score. Here is a detailed explanation of the model layers.

\subsubsection{Input layer}
This layer delineates the configuration of the input data that the model will process. It anticipates each sample to possess a shape of $(X\_train.shape[1], 1)$, where the initial entry denotes the number of features in the input data, which is 15 in this instance. The numeral 1 signifies that the input data is one-dimensional, such as a time series or a sequence of features. This is the initial phase of the model, ensuring that the input data is appropriately formatted. The input data can be represented as $ [X \in {\mathbb{R}^{15 \times 1}}]$.
\subsubsection{Conv1D layer}
The input data is subjected to 16 filters (small learnable weights) by the first Conv1D layer. Because each filter has a kernel size of 3, it can examine three features in succession at once. The model learns the spatial correlations between features with the aid of this layer. For example, it could discover patterns in an educational dataset connecting students' activity data over time or between various kinds of interactions (such forum postings and assignments). The activation function $ReLU$ adds non-linearity to the model, enabling it to learn intricate patterns. In order to help avoid negative numbers from distorting the results, $ReLU$ outputs zero if the value is not positive. If the value is positive, it outputs the value directly. The convolution operation in mathematics is expressed as;
\begin{equation}
{Y_{1CL}}[i,k] = ReLU\left( {\sum\limits_{j = 0}^3 {{W_{1,k,j}}.X[i + j - 1] + {b_{1,k}}} } \right),
\end{equation}

where ${Y_{1CL}}[i,k]$ is the output for filter $k$ at position $i$, ${{W_{1,k,j}}}$ are the filter weights, and ${{b_{1,k}}}$ is the bias for filter $k$. The shape of output after applying first 1D-convolutional operation is ${Y_{1CL}} \in {\mathbb{R}^{15 \times 16}}$.
\subsubsection{Maxpooling layer}
This layer does down-sampling by selecting the maximum value within each 2-feature frame, hence halving the data's dimensionality (from 15 to 7). Max pooling decreases data complexity by emphasising the most salient characteristics identified by the convolutional layer. It also aids in rendering the model invariant to minor translations of the input (e.g., modest fluctuations in feature values). It can be expressed as; 
\begin{equation}
{Z_{1MP}}[i,k] = \max ({Y_{1CL}}[2i,k],{Y_{1CL}}[2i + 1,k]),
\end{equation}
where ${Z_{1MP}}[i,k]$ is the down-sampled output for the $ith$ position and $kth$ filter. The shape of data after max pooling is ${Z_{1MP}}\in {\mathbb{R}^{7 \times 16}}$.
\subsubsection{Batch normalization layer}
Batch normalization standardizes the outputs of preceding layers. It modifies and normalizes the activation by ensuring that the mean of each batch approximates 0 and the standard deviation approaches 1. This layer stabilises the training process by mitigating significant fluctuations in data distribution as it traverses the network, hence facilitating faster convergence and enhancing generalisation.
\begin{equation}
{{\mathord{\buildrel{\lower3pt\hbox{$\scriptscriptstyle\frown$}} 
\over Y} }_{BN}}[i,k] = \frac{{{Y_{6CL}}[i,k] - {\mu _k}}}{{\sqrt {\sigma _k^2 + \varepsilon } }},
\end{equation}
where ${\mu _k}$, ${\sigma _k^2}$, and $\varepsilon $ is the mean of $kth$ feature map, variance of $kth$ feature map, and the small constant added for numerical stability, respectively.

The multi-dimensional data from the convolutional layers is flattened into a one-dimensional (1D) vector by the flatten layer. To feed the data into the dense (fully connected) layers, it must first be flattened. It converts the 3D tensor (batch size, feature-length, number of filters) into a 2D tensor (batch size, total features).
\begin{equation}
   {Y_{FL}} = \varsigma ({{\mathord{\buildrel{\lower3pt\hbox{$\scriptscriptstyle\frown$}} 
\over Y} }_{BN}}),
\end{equation}
where $\varsigma$ represents the flatten operation, ${{\mathord{\buildrel{\lower3pt\hbox{$\scriptscriptstyle\frown$}} 
\over Y} }_{BN}}$ is the output of the batch normalization, and ${Y_{FL}}$ is the output of flatten layer.
\subsubsection{Fully connected layers} 
The initial fully connected (dense) layer comprises 128 neurons. Every neuron is entirely linked to all outputs from the preceding layer.
Dense layers execute the classification operation by analysing the characteristics derived from the convolutional layers. The $ReLU$ activation facilitates the acquisition of intricate correlations among the characteristics. Mathematically represented as 
\begin{equation}
{Z_{1DL}} = ReLU({W_{1DL}}.{Y_{FL}} + {b_{1DL}})
\end{equation}

The output dimension of the initial fully connected layer is ${Z_{1DL}} \in {\mathbb{R}^{128}}$. 
The second dense layer comprises 64 neurons, which further distils the information, enabling the model to concentrate on the most salient aspects. This layer further refines the features for the ultimate classification. 
The third dense layer comprises 32 neurons, substantially diminishing the data's dimensionality and approaching the number of output classes. This further refines the model's focus, aiding in the elimination of extraneous features and retaining only those most important for classification.
The penultimate dense layer preceding the output layer comprises 16 neurons, facilitating data preparation for the final classification phase. This layer guarantees that the data is appropriately formatted and sized for the final classification phase. The operation of Equation (5) has been executed for the second, third, and fourth dense layers. The data size following the fourth dense layer is ${Z_{4DL}} \in {\mathbb{R}^{16}}$.
\subsubsection{Dropout layer}
During each training step, the Dropout layer randomly sets $30\%$ of the input units to 0, denoted as ${Z_{drop}} = ({Z_{4DL}},\rho = 0.3)$. By keeping the model from being overly reliant on any one neuron, this keeps it from overfitting. Dropout increases the model's capacity for generalisation and reduces the likelihood that it will overfit the training set.
 \subsubsection{Output layer}
 There are $n\_classes=4$ units in the output layer, one for each class. A probability distribution over the classes is created by using the softmax activation function, where each unit denotes the likelihood that the input belongs to a certain class. Multi-class classification issues are solved using softmax, which is expressed as ${Z_{final}} = soft\max ({Z_{drop}})$. Data ${Z_{final}} \in {\mathbb{R}^4}$ is the size of the data. This layer outputs the probability distribution among the classes. This layer outputs the probability distribution among the classes. The reason for using sparse categorical cross-entropy is that the labels are integers rather than one-hot encoded. For multi-class classification problems, this loss function is suitable, as it is described as
\begin{equation}
loss =  - \frac{1}{N}\sum\limits_{i = 1}^N {\log (P({y_i}|{X_i}))}, 
\end{equation}

where $N$ denotes the number of samples and $(P({y_i}|{X_i}))$ represents the projected probability for the true class $y_i$ given the input $X_i$. Adam serves as the optimiser. Adam is an optimisation method with an adaptive learning rate that has gained significant popularity owing to its efficiency and performance.

This one-dimensional (1D) model analyses sequential data by employing convolutional layers for feature extraction, downsampling via max pooling, and normalizing through batch normalization. The collected characteristics are further flattened and transmitted across many fully connected layers, incorporating dropout to mitigate overfitting. Ultimately, the output undergoes a softmax function to produce a probability distribution among the four classes. The model employs sparse categorical cross-entropy as the loss function and is optimised with the Adam optimiser. 

\section{Experimental Results and Discussion}
This section presents the experimental setup, results, and discussion. We use the data pre-processed dataset whose details are mentioned in Section \ref{sec:sample3}.
\subsection{Environment setup}
For the implementation of baseline and proposed models, Jupyter Notebook was utilized where required libraries such as Pandas, Seaborn, Numpy, Keras, and TensorFlow were imported. The experiments were conducted on the Kaggle with the GPU support.

\subsection{Experimental results}
\iffalse
\begin{figure}[h]
	\centering
	% \resizebox{.49\textwidth}{.17\textheight}{%
		\includegraphics[width=0.6\linewidth]{Final Grade Class Samples.PNG}%}
	\caption{Final Grade Class Samples. }
        \label{fig:fig6 final grade class samples}
\end{figure}

Figure \ref{fig:fig6 final grade class samples} illustrates the distribution of samples across four distinct final grade categories: Pass, Distinction, Fail, and Withdrawn. The bulk of samples are classified in the Pass category, comprising $59.485\%$ of the total. The subsequent prevalent category is Distinction, with $17.982\%$ of samples. The Fail and Withdrawn categories comprise $13.227\%$ and $9.306\%$ of the samples, respectively. The graphic indicates that the majority of students in the dataset have successfully passed, while a lesser fraction attained distinction, failing, or withdrawing from their courses.
\fi

\begin{table*}[h]
	\caption{Comparison of proposed method with RF 'gini' and RF 'entropy' training with all parameters.} 
	\setlength{\tabcolsep}{3pt} % Adjust column separation
	\resizebox{\textwidth}{!}{% Resize table to text width
		\begin{tabular}{l lllllllllll}
			\toprule
			\multicolumn{1}{l}{} & \multicolumn{3}{c}{Proposed} & & \multicolumn{3}{c} {RF 'gini' \cite{b15}} & & \multicolumn{3}{c} {RF 'entropy' \cite{b15}}\\ 
			\cmidrule(lr){2-4} \cmidrule(lr){6-8} \cmidrule(lr){10-12}
			& Precision & Recall & F1-score & & Precision & Recall & F1-score & & Precision & Recall & F1-score \\
			\midrule
            Distinction & \textbf{0.97} & \textbf{0.99} & \textbf{0.98} & & 0.66 & 0.48 & 0.56 & & 0.66 & 0.48 & 0.56  \\
            Fail & \textbf{0.97} & \textbf{0.96} & \textbf{0.96} & & 0.56 & 0.45 & 0.50 & & 0.55 & 0.44 & 0.49  \\
            Pass & \textbf{0.99} & \textbf{0.99} & \textbf{0.99} & & 0.69 & 0.86 & 0.77 & & 0.69 & 0.87 & 0.77 \\
            Withdrawn & \textbf{1.00} & \textbf{0.99} & \textbf{0.99} & & 0.64 & 0.50 & 0.57 & & 0.65 & 0.48 & 0.56  \\
            Accuracy & \textbf{0.98} &  &  & & 0.66 &  &  & & 0.66 &  &    \\    
            Macro average & \textbf{0.98} & \textbf{0.98}  & \textbf{0.98} & & 0.64 & 0.57  & 0.60  & & 0.64 & 0.57 &  0.59  \\  
            Weighted average & \textbf{0.98} & \textbf{0.98}  & \textbf{0.98}  & & 0.65 & 0.66  & 0.65  & & 0.65 & 0.66 &  0.64   \\  
            \bottomrule
	    \end{tabular}% End of tabular
    }% End of resizebox
    \label{tab:tab1 compared with RF}
\end{table*}

Table \ref{tab:tab1 compared with RF} compares the efficacy of the proposed strategy against RFs models utilising 'gini' and 'entropy' criteria \cite{b15}, trained and evaluated on the comprehensive OULAD dataset for multiclass classification (Distinction, Pass, Fail, and Withdrawn categories). The assessment measures encompass accuracy, precision, recall, and F1 score for all four categories. The suggested strategy regularly surpasses the baseline RF 'gini' and RF 'entropy' models, yielding improved outcomes across all metrics. This suggests that the suggested strategy is superior in accurately identifying student outcomes, particularly in achieving a balance between precision and recall, resulting in more dependable and robust predictions. The improved F1 ratings further emphasise its capacity to adeptly balance precision and recall, rendering it a more potent instrument for forecasting student performance across various categories.

\iffalse 
%%% Table 2
\begin{table*}[h]
	\caption{Comparison of the proposed method with the DFFNN \cite{b10} method when using only demographic data.} 
	\setlength{\tabcolsep}{3pt}
	\resizebox{\textwidth}{!}{%
		\begin{tabular}{l lllllll}
			\toprule
			\multicolumn{1}{l}{} & \multicolumn{3}{c}{Proposed} & & \multicolumn{3}{c}{DFFNN \cite{b10}} \\ 
			\cmidrule(lr){2-4} \cmidrule(lr){6-8}
			& Precision & Recall & F1-score & & Precision & Recall & F1-score \\
			\midrule
                Distinction & \textbf{0.89} & \textbf{0.97} & \textbf{0.93} & & 0.56 & 0.01 & 0.02 \\
                Fail & \textbf{0.88} & \textbf{0.88} & \textbf{0.88} & & 0.39 & 0.12 & 0.18 \\
                Pass & \textbf{0.95} & \textbf{0.96} & \textbf{0.95} & & 0.44 & 0.70 & 0.54 \\
                Withdrawn & \textbf{0.94} & \textbf{0.77} & \textbf{0.85} & & 0.42 & 0.45 & 0.44 \\
                Accuracy & \textbf{0.93} &  &  & & 0.43 &  &  \\    
                Macro average & \textbf{0.92} & \textbf{0.89}  & \textbf{0.90} & & 0.45 & 0.32 & 0.30 \\  
                Weighted average & \textbf{0.93} & \textbf{0.93}  & \textbf{0.93} & & 0.43 & 0.43 & 0.38 \\  
            \bottomrule
	    \end{tabular}%
    }% End of resizebox
    \label{tab:tab2 Demographic}
\end{table*}
%Table 2
\fi

\begin{table*}[h]
	\caption{Comparison of the proposed method with the DFFNN \cite{b10} and ANN-LSTM \cite{b20} when using demographic and clickstream data.} 
	\setlength{\tabcolsep}{3pt} % Adjust column separation
	\resizebox{\textwidth}{!}{%
		\begin{tabular}{l lllllllllll}
			\toprule
			\multicolumn{1}{l}{} & \multicolumn{3}{c}{Proposed} & & \multicolumn{3}{c}{ANN-LSTM \cite{b20}} & & \multicolumn{3}{c}{DFFNN \cite{b10}} \\ 
			\cmidrule(lr){2-4} \cmidrule(lr){6-8} \cmidrule(lr){10-12}
			& Precision & Recall & F1-score & & Precision & Recall & F1-score & & Precision & Recall & F1-score \\
			\midrule
                Distinction & \textbf{0.92} & \textbf{0.97} & \textbf{0.94} & & 0.82 & 0.47 & 0.59 & & 0.77 & 0.02 & 0.04 \\
                Fail & \textbf{0.88} & \textbf{0.90} & \textbf{0.89} & & 0.79 & 0.55 & 0.65 & & 0.52 & 0.20 & 0.29 \\
                Pass & \textbf{0.96} & \textbf{0.96} & \textbf{0.96} & & 0.84 & 0.60 & 0.70 & & 0.62 & 0.92 & 0.74 \\
                Withdrawn & \textbf{0.93} & \textbf{0.80} & \textbf{0.86} & & 0.82 & 0.62 & 0.70 & & 0.67 & 0.76 & 0.71 \\
                Accuracy & \textbf{0.94} &  &  & & 0.72 &  &  & & 0.63 &  & \\    
                Macro average & \textbf{0.92} & \textbf{0.91} & \textbf{0.91} & & 0.82 & 0.56 & 0.66 & & 0.64 & 0.48 & 0.45 \\  
                Weighted average & \textbf{0.94} & \textbf{0.94} & \textbf{0.94} & & 0.82 & 0.58 & 0.68 & & 0.63 & 0.63 & 0.56 \\  
            \bottomrule
	    \end{tabular}%
    }% End of resizebox
    \label{tab:tab3 Compared with ANN-LSTM and DFFNN}
\end{table*}

Table \ref{tab:tab3 Compared with ANN-LSTM and DFFNN} provides a comparative study of the suggested method in relation to the ANN-LSTM \cite{b20} and DFFNN \cite{b10} models, assessed on the OULAD dataset utilising demographic and clickstream variables for multiclass classification (Distinction, Pass, Fail, and Withdrawn categories). The suggested model exhibits enhanced performance across all assessment criteria, including accuracy, precision, recall, and F1 score. Its superior accuracy signifies that it generates a greater number of right predictions overall. The increased accuracy indicates that when the model forecasts a particular result, such as Distinction, it is more probable to be accurate, thereby diminishing the frequency of false positives. The enhanced recall values indicate that the model is more adept at identifying all true occurrences of each class, so successfully reducing false negatives. The F1 score, which equilibrates precision and recall, is continuously superior for the suggested method, underscoring its capacity to navigate the trade-off between these two criteria more adeptly than the baseline models. The overall performance enhancement indicates that the suggested method is especially proficient in managing the intricacies of student performance prediction, providing more dependable and nuanced insights across all categories of student outcomes.

\begin{table*}[h]
	\caption{Comparison of the proposed method with the DFFNN \cite{b10} when using demographic, clickstream, and assessment data.} 
	\setlength{\tabcolsep}{3pt}
	\resizebox{\textwidth}{!}{%
		\begin{tabular}{l lllllll}
			\toprule
			\multicolumn{1}{l}{} & \multicolumn{3}{c}{Proposed} & & \multicolumn{3}{c}{DFFNN \cite{b10}} \\ 
			\cmidrule(lr){2-4} \cmidrule(lr){6-8}
			 & Precision & Recall & F1-score & & Precision & Recall & F1-score \\
			\midrule
                Distinction & \textbf{0.91} & \textbf{0.98} & \textbf{0.95} & & 0.66 & 0.47 & 0.55 \\
                Fail & \textbf{0.92} & \textbf{0.90} & \textbf{0.91} & & 0.55 & 0.36 & 0.44 \\
                Pass & \textbf{0.96} & \textbf{0.97} & \textbf{0.97} & & 0.74 & 0.90 & 0.81 \\
                Withdrawn & \textbf{0.95} & \textbf{0.83} & \textbf{0.89} & & 0.76 & 0.80 & 0.78 \\
                Accuracy & \textbf{0.95} &  &  & & 0.71 &  & \\    
                Macro average & \textbf{0.94} & \textbf{0.92} & \textbf{0.93} & & 0.68 & 0.63 & 0.64 \\  
                Weighted average & \textbf{0.95} & \textbf{0.95} & \textbf{0.95} & & 0.70 & 0.71 & 0.70 \\  
            \bottomrule
	    \end{tabular}%
    }% End of resizebox
    \label{tab:tab4 Demographic clickstream and assessment}
\end{table*}

\begin{table*}[h]
	\caption{Comparison of the proposed method with the DFFNN \cite{b10} when using all variables for training and testing.} 
	\setlength{\tabcolsep}{3pt}
	\resizebox{\textwidth}{!}{%
		\begin{tabular}{l lllllll}
			\toprule
			\multicolumn{1}{l}{} & \multicolumn{3}{c}{Proposed} & & \multicolumn{3}{c}{DFFNN \cite{b10}} \\ 
			\cmidrule(lr){2-4} \cmidrule(lr){6-8}
			& Precision & Recall & F1-score & & Precision & Recall & F1-score \\
			\midrule
                Distinction & \textbf{0.97} & \textbf{0.99} & \textbf{0.98} & & 0.64 & 0.49 & 0.55 \\
                Fail & \textbf{0.97} & \textbf{0.96} & \textbf{0.96} & & 0.55 & 0.37 & 0.44 \\
                Pass & \textbf{0.99} & \textbf{0.99} & \textbf{0.99} & & 0.76 & 0.91 & 0.82 \\
                Withdrawn & \textbf{1.00} & \textbf{0.99} & \textbf{0.99} & & 0.76 & 0.80 & 0.78 \\
                Accuracy & \textbf{0.98} &  &  & & 0.72 &  & \\    
                Macro average & \textbf{0.98} & \textbf{0.98} & \textbf{0.98} & & 0.68 & 0.64 & 0.65 \\  
                Weighted average & \textbf{0.98} & \textbf{0.98} & \textbf{0.98} & & 0.70 & 0.72 & 0.70 \\  
            \bottomrule
	    \end{tabular}%
    }% End of resizebox
    \label{tab:tab5 complete dataset}
\end{table*}

Three distinct scenarios are used to compare the proposed method with DFFNN \cite{b10}: using only demographic features; combining clickstream, assessment, and demographic features; and finally using the entire dataset with all features as shown in Table \ref{tab:tab4 Demographic clickstream and assessment}, and Table \ref{tab:tab5 complete dataset}, respectively. For the four classes of Distinction, Pass, Fail, and Withdrawn, the results are displayed in terms of accuracy, precision, recall, F1-score, macro average, and weighted average. The suggested technique consistently performs better than the DFFNN in all tables, especially when more features are taken into account. This pattern suggests that the suggested approach performs better at using the combined features to enhance classification performance. Greater precision and recall suggest a more equitable treatment of true positives and false positives across all classes, while higher accuracy points to better overall predictions. The improved F1-score, macro, and weighted averages demonstrate the robustness of the suggested approach and demonstrate its capacity to generate accurate and nuanced forecasts in a range of educational contexts.

\begin{figure}[h]
	\centering
	% \resizebox{.49\textwidth}{.17\textheight}{%
		\includegraphics[width=0.6\linewidth]{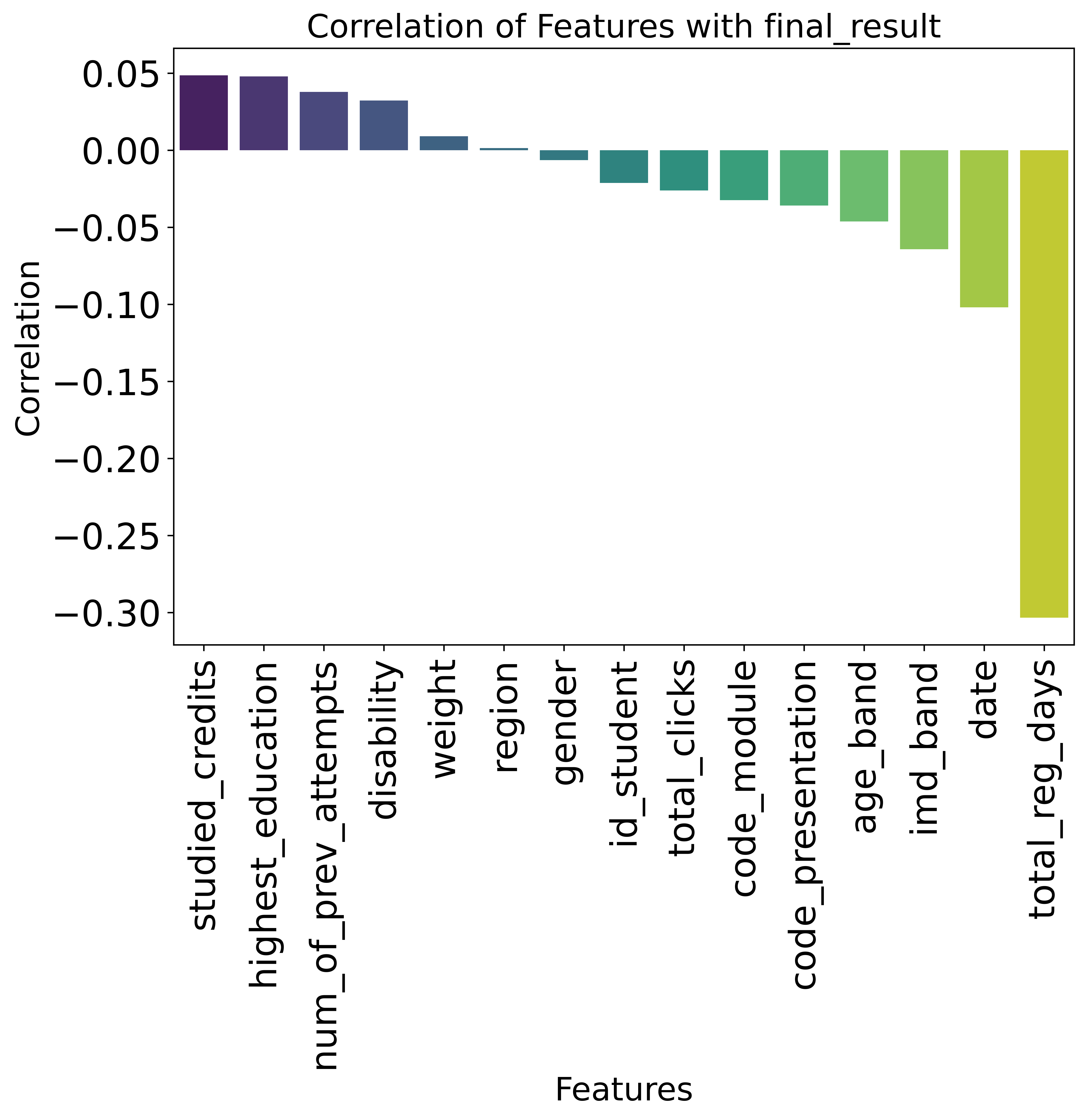}%}
	\caption{Correlation analysis of features with final\_result. }
        \label{fig:fig6 correlation matrix}
\end{figure}

Figure \ref{fig:fig6 correlation matrix} depicts the influence of several factors on the final output. The $total\_reg\_days$ is identified as the most relevant feature, with a significance value of $-0.30$. Other features also influence performance, including $studied\_credits$, $highest\_education$, $imd\_band$, and $num\_of\_prev\_attempts$, among others. The $total\_reg\_days$ feature is derived through feature engineering from the $reg\_date$ and $unreg\_date$. Only the $unreg\_date$ for the Withdrawn class is accessible in the OULAD. The $unreg\_date$ is crucial for other classes for differentiation, as evidenced by the value of $total\_reg\_days$ illustrated in the Figure \ref{fig:fig6 correlation matrix}.

\begin{figure}[h]
	\centering
	% \resizebox{.49\textwidth}{.17\textheight}{%
		\includegraphics[width=0.6\linewidth]{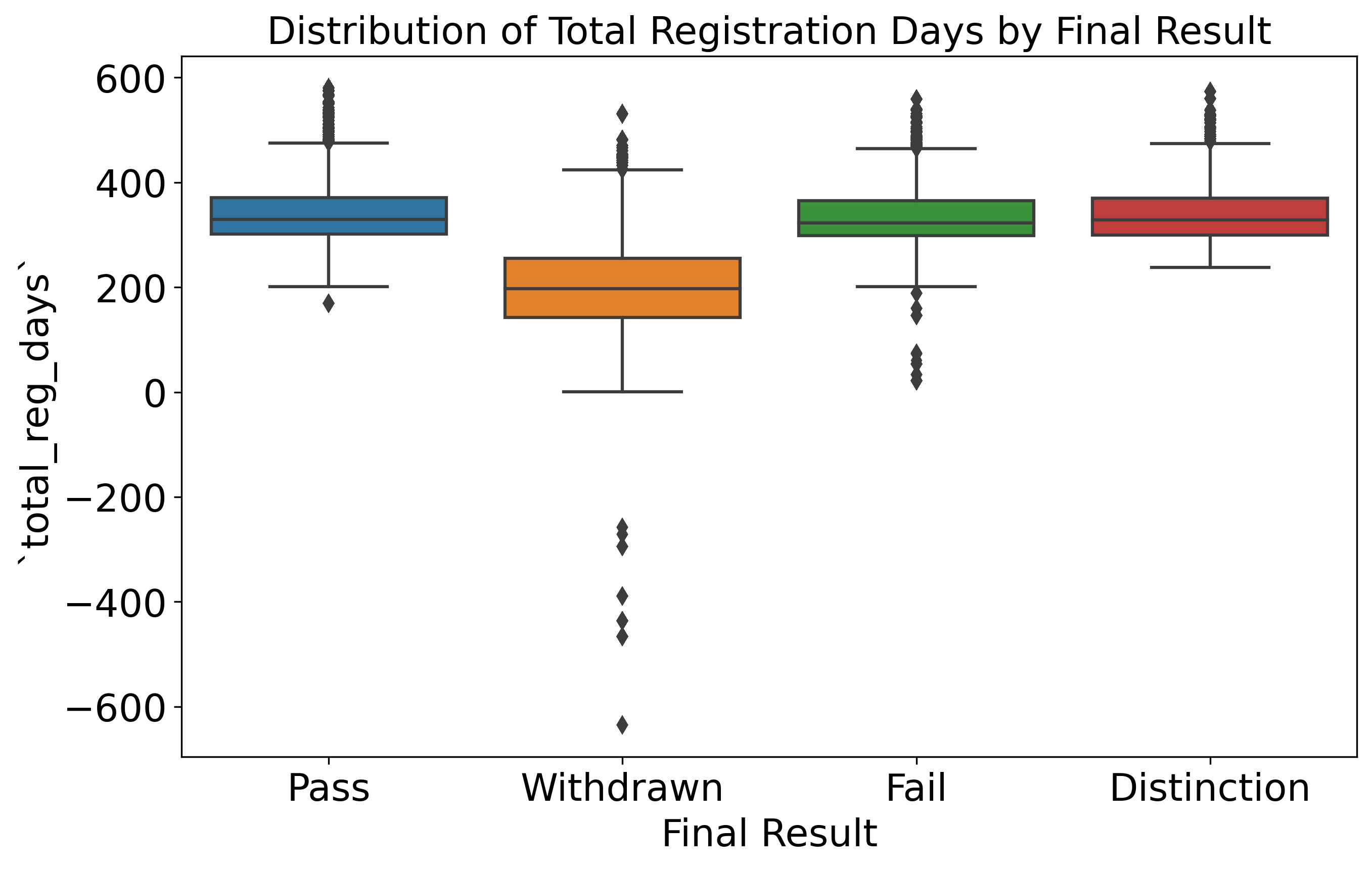}%}
	\caption{Class distribution against $total\_reg\_days$. }
        \label{fig:fig7 class duration}
\end{figure}

Figure \ref{fig:fig7 class duration} shows the effect of feature $total\_reg\_days$ on the Final Results of the students, which is created by using feature engineering on the features $registration\_date$ and $unregistration\_date$ (only mentioned of Withdrawn class in the dataset). This feature clearly represents the discrimination in all classes including Distinction, Fail, Pass, and Withdrawn.

\subsection{Early Prediction}

\begin{figure}[h]
	\centering
	\subfigure[Course duration $5\%$]{\includegraphics[width=0.45\linewidth]{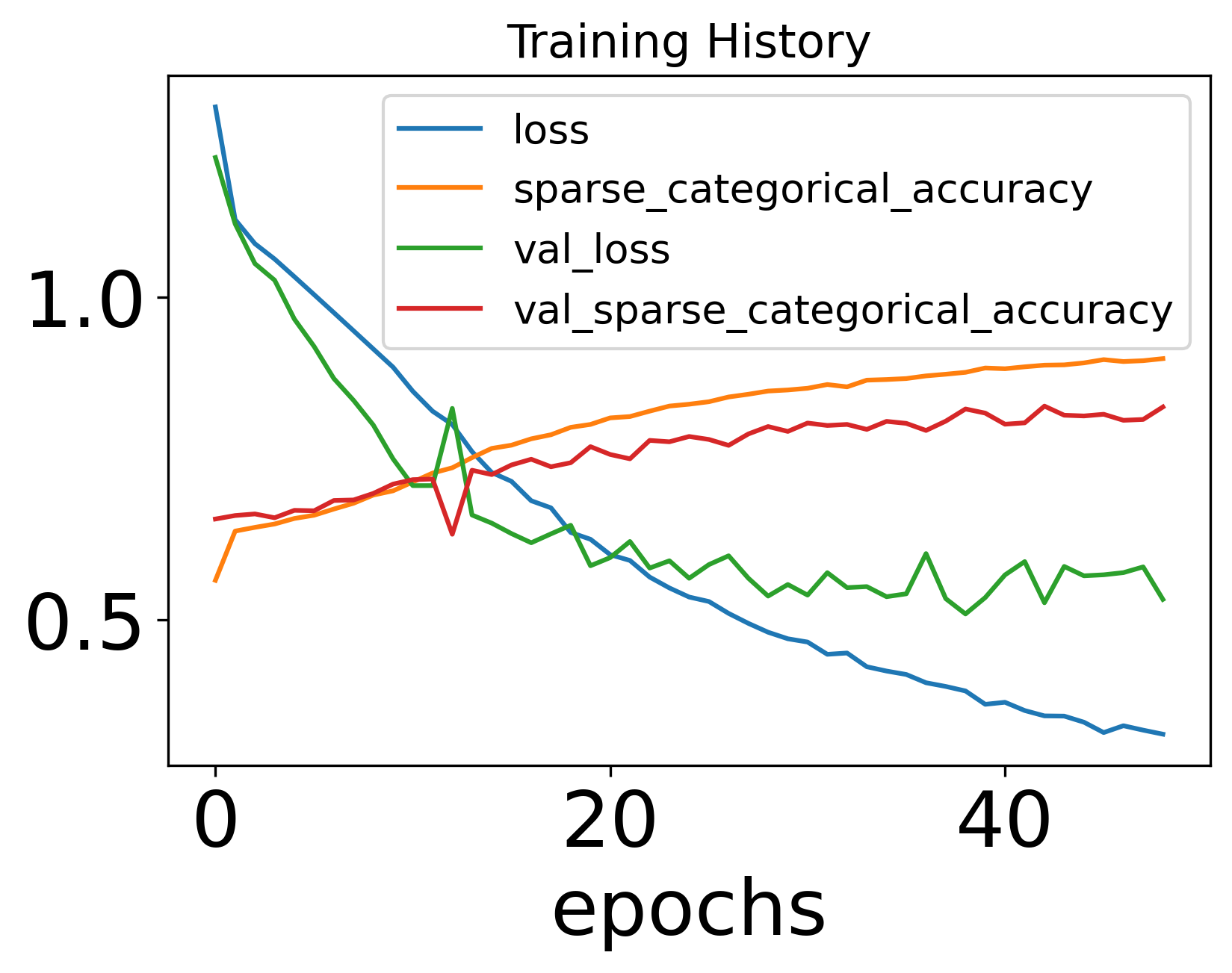}}
	\subfigure[$10\%$ of dataset]{\includegraphics[width=0.45\linewidth]{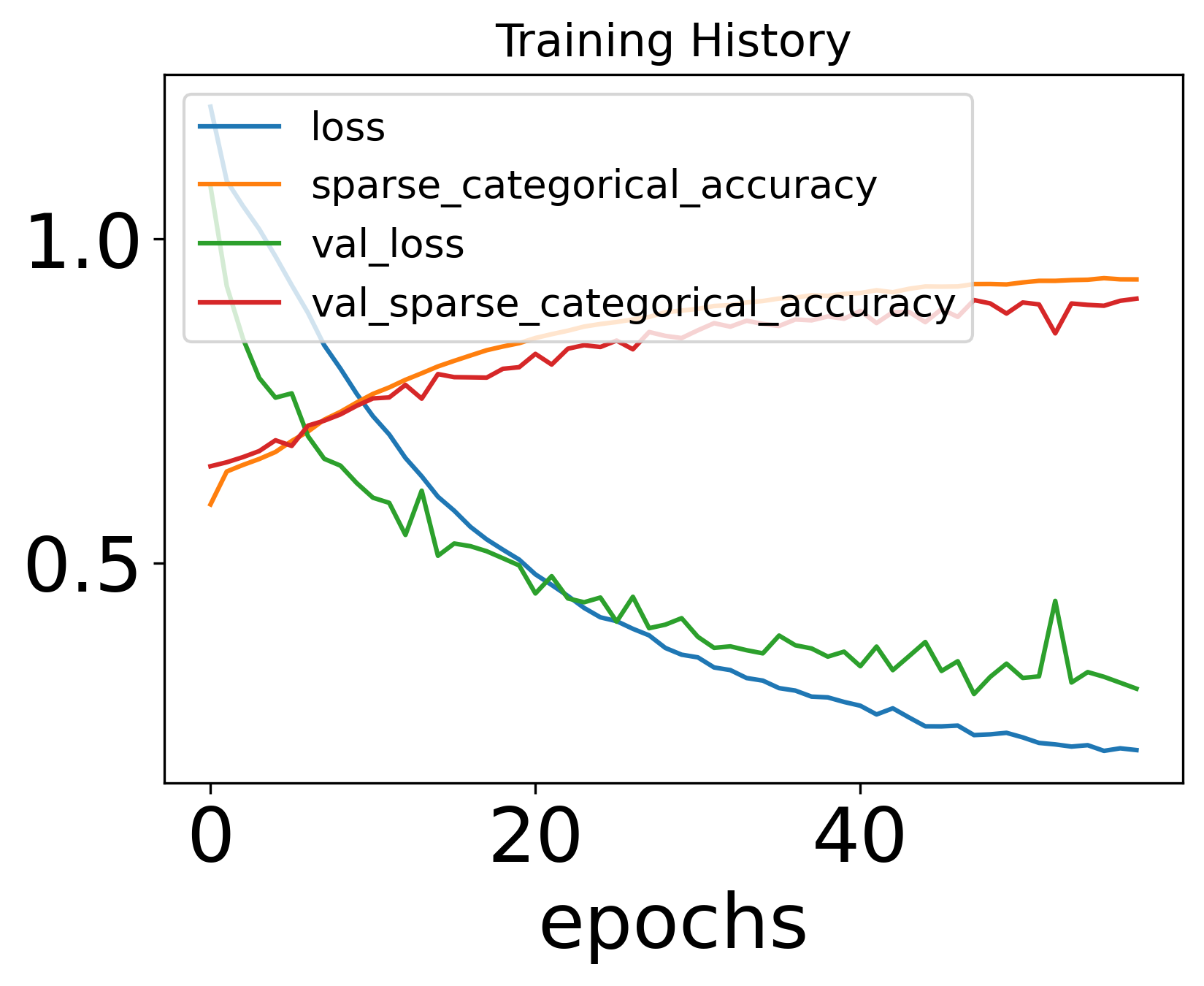}}
	\subfigure[Course duration $20\%$]{\includegraphics[width=0.45\linewidth]{Training_History_CNN_10_Course_new.png}}
 	%\subfigure[$40\%$ of dataset]{\includegraphics[width=0.32\linewidth]{Training History CNN 40 Course_new.png}}
  	%\subfigure[$60\%$ of dataset]{\includegraphics[width=0.32\linewidth]{Training History CNN 60 Course_new.png}}
   	%\subfigure[Course duration $80\%$]{\includegraphics[width=0.40\linewidth]{Training History CNN 80 Course.png}}
    	\subfigure[Course duration $100\%$]{\includegraphics[width=0.45\linewidth]{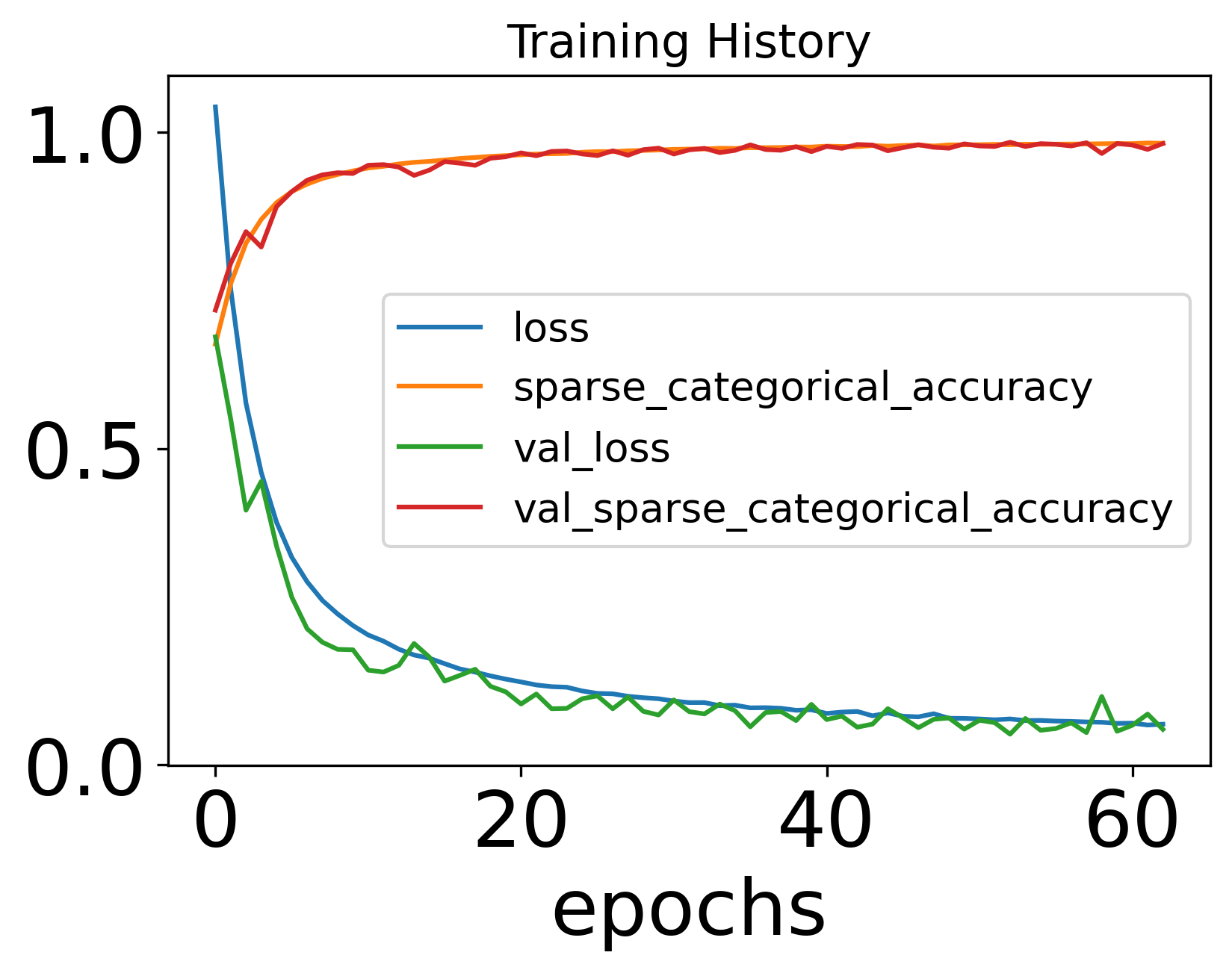}}
	\caption{Training vs validation accuracy and loss at different training of dataset size  (a) $5\%$ (b) $10\%$ (c) $20\%$ and (d) $100\%$.}
	\label{fig:fig5 training and validation accuracy and loss}
\end{figure}

Figure \ref{fig:fig5 training and validation accuracy and loss} compares training and validation accuracy as well as loss to show how varying course length percentages ($5\%$, $10\%$, $20\%$, and $100\%$) affect model performance. The model exhibits better generalisation as the dataset size grows, with a more steady validation loss and a lower difference between training and validation accuracy. Significant overfitting occurs when the model performs well on training data but badly on validation data, and this is the case with smaller datasets (e.g., $5\%$). The training and validation measures, on the other hand, more closely correlate with $100\%$ of the dataset, suggesting that the model benefits from more data and is less prone to overfitting. This illustrates how higher model performance and generalisation can be attained with larger datasets. Even when trained at a small dataset size of $20\%$, our suggested model achieves improved training and validation accuracy as well as loss. When the data size increases from $5\%$ to $10\%$ and $20\%$, the model flawlessly aids in the early prediction of students who are at risk.

\begin{figure}[h]
	\centering
	\subfigure[Accuracy]{\includegraphics[width=0.45\linewidth]{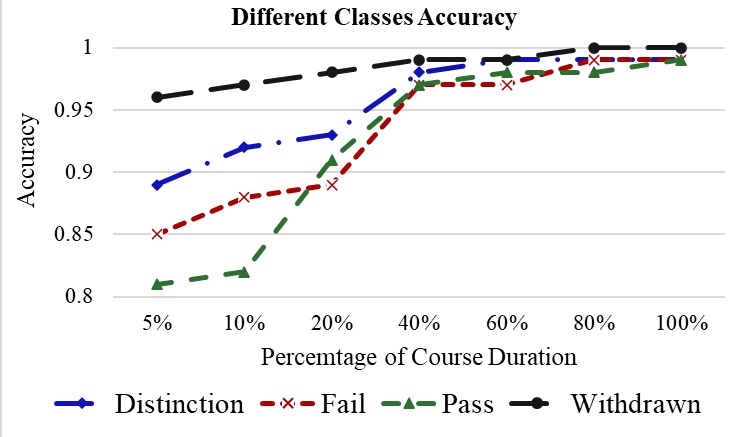}}
	\subfigure[Precision]{\includegraphics[width=0.45\linewidth]{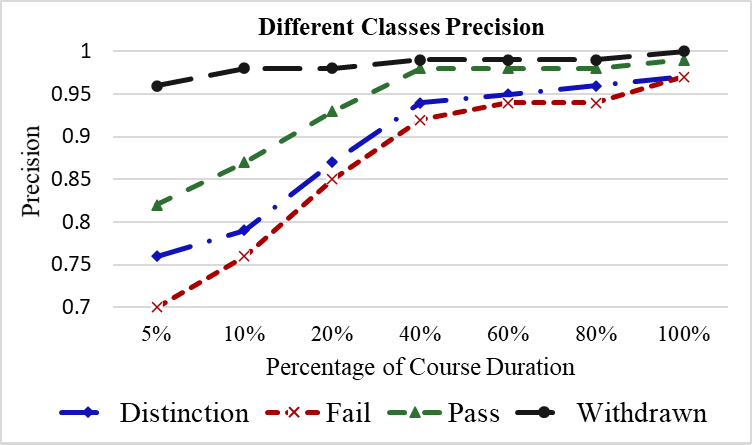}}
   	\subfigure[Recall]{\includegraphics[width=0.45\linewidth]{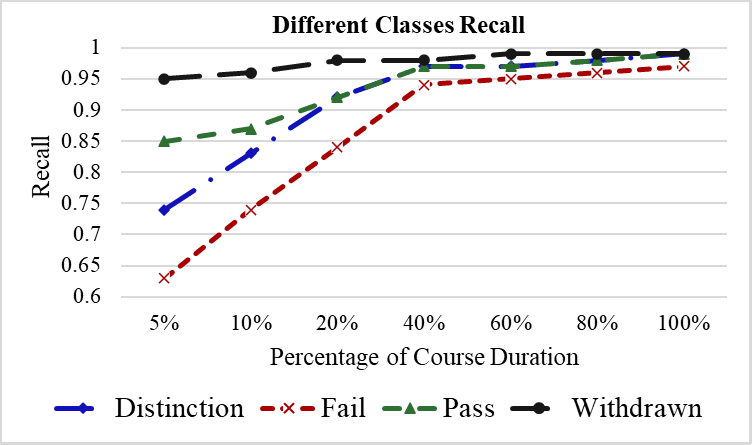}}
    	\subfigure[F1-score]{\includegraphics[width=0.45\linewidth]{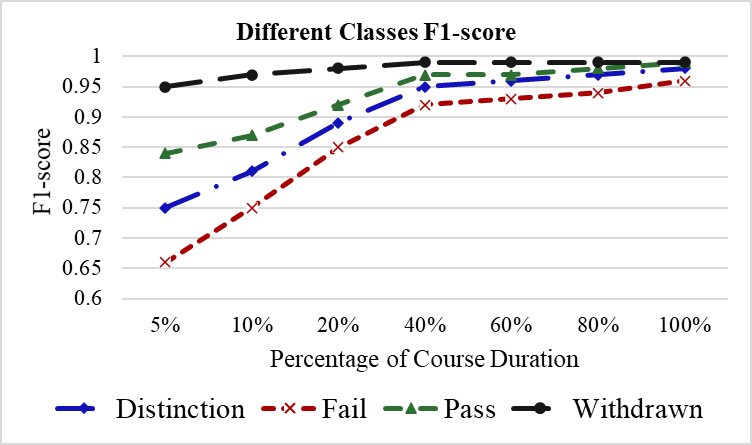}}
	\caption{Performance of different classes with the percentage of course length (a) Accuracy (b) Precision (c) Recall and (d) F1-score.}
	\label{fig:fig5 accuracy precision-recall and f1}
\end{figure}
Figure \ref{fig:fig5 accuracy precision-recall and f1} shows the model performance of accuracy, precision, recall, and F1-score while considering the $5\%$, $10\%$, $20\%$, $40\%$, $60\%$, $80\%$ and $100\%$ course duration. The results of classes: Distinction, Fail, and Pass are more than $90\%$ in terms of accuracy, precision, recall, and F1-score as the course duration increases more than $20\%$. While the Withdrawn class performance is much better than other classes even at $5\%$ of course duration. The reason is that the $reg\_date$ and $unreg\_date$ of Withdrawn students are mentioned but for other classes students only $reg\_date$ is available in the OULAD dataset. Thus, we have created a new feature $total\_reg\_days= reg\_date-unreg\_date$ where a value of 270 has been considered for classes whose $unreg\_date$ is not available in the dataset because of the last day of course duration.

\begin{table*}[h!]
	\caption{Early prediction of students' performance in online learning while considering all features.} 
	\setlength{\tabcolsep}{3pt}
	\resizebox{\textwidth}{!}{%
		\begin{tabular}{l lllllllllll}
			\toprule
			\multicolumn{1}{l}{} & \multicolumn{3}{c}{$5\%$ course duration} && \multicolumn{3}{c}{$10\%$ course duration} & &\multicolumn{3}{c}{$20\%$ course duration}\\ 
			\cmidrule(lr){2-4} \cmidrule(lr){6-8} \cmidrule(lr){10-12}
			 & Precision & Recall & F1-score & & Precision & Recall & F1-score & & Precision & Recall & F1-score \\
			\midrule
                Distinction & 0.76 & 0.74 & 0.75 & & 0.79 & 0.83 & 0.81 & & 0.87 & 0.92 & 0.89\\
                Fail & 0.70 & 0.63 & 0.66 & & 0.76 & 0.74 & 0.75 & & 0.85 & 0.84 & 0.85\\
                Pass & 0.82 & 0.85 & 0.84 & & 0.87 & 0.87 & 0.87 & & 0.93 & 0.92 & 0.92\\
                Withdrawn & 0.96 & 0.95 & 0.95 & & 0.98 & 0.96 & 0.97 & & 0.98 & 0.98 & 0.98\\
                Accuracy & 0.82 &  &  & & 0.86 &  &  & & 0.92 &  & \\    
                Macro average & 0.81 & 0.79 & 0.80 & & 0.85 & 0.85 & 0.85 & & 0.91 & 0.91 & 0.91 \\  
                Weighted average & 0.82 & 0.82 & 0.82 & & 0.86 & 0.86 & 0.86 & & 0.92 & 0.92 & 0.92\\  
                \midrule
                
                \multicolumn{1}{l}{} & \multicolumn{3}{c}{$40\%$ course duration} & & \multicolumn{3}{c}{$60\%$ course duration} & & \multicolumn{3}{c}{$100\%$ course duration}\\ 
			\cmidrule(lr){2-4} \cmidrule(lr){6-8} \cmidrule(lr){10-12} 
  
   			 & Precision & Recall & F1-score & & Precision & Recall & F1-score & & Precision & Recall & F1-score \\
        \midrule
                Distinction  & 0.94 & 0.97 & 0.95 & & 0.95 & 0.97 & 0.96 & & \textbf{0.97} & \textbf{0.99} & \textbf{0.98}\\
                Fail  & 0.94 & 0.94 & 0.94 & & 0.92 & 0.95 & 0.93 & & \textbf{0.97} & \textbf{0.96} & \textbf{0.96}\\
                Pass  & 0.98 & 0.97 & 0.97 & & 0.98 & 0.97 & 0.97 & & \textbf{0.99} & \textbf{0.99} & \textbf{0.99}\\
                Withdrawn  & 0.99 & 0.98 & \textbf{0.99} & & \textbf{1.00} & \textbf{0.99} & \textbf{0.99} & & \textbf{1.00} & \textbf{0.99} & \textbf{0.99}\\
                Accuracy  & 0.97 &  &  & & 0.97 &  &  & & \textbf{0.98} &  & \\    
                Macro average  & 0.96 & 0.96 & 0.96 & & 0.96 & 0.97 & 0.96 & & \textbf{0.98} & \textbf{0.98} & \textbf{0.98}\\  
                Weighted average  & 0.97 & 0.97 & 0.97 & & 0.97 & 0.97 & 0.97 & & \textbf{0.98} & \textbf{0.98} & \textbf{0.98}\\  
            \bottomrule
	\end{tabular}}  
    \label{tab:tab6 early prediction of performance while using all features}
\end{table*}

Taking into account all relevant features, Table \ref{tab:tab6 early prediction of performance while using all features} shows the early prediction performance of students in online learning at different points in the course progression. Metrics like accuracy, precision, recall, and F1-score are used to assess performance at various prediction intervals including $5\%$, $10\%$, $20\%$, $40\%$, $60\%$, and $100\%$ of the course duration. The following categories are evaluated: Withdrawn, Distinction, Fail, and Pass. When it comes to Distinction, the model performs moderately at first (Precision: $0.76$, Recall: $0.74$, F1-score: $0.75$) at $5\%$ of course duration, but by the end of the course, it has much improved ($0.97$, $0.99$, $0.98$), demonstrating high accuracy in identifying students who are likely to excel. For Fail class, predictions start off with lower scores ($0.70$, $0.63$, $0.66$) at $5\%$ of course duration, but these rise to $0.97$, $0.96$, and $0.96$ by considering $100\%$ of course duration, demonstrating that more data leads to a better identification of failed students. For Pass, the model shows dependable forecasts of passing students early on ($0.82$, $0.85$, $0.84$) and reaches near-perfect accuracy ($0.99$ for all metrics) at $100\%$ of course length. Withdrawn exhibits the best overall performance, attaining near-perfect scores by $100\%$ course duration ($1.00$, $0.99$, $0.99$) after beginning high at $5\%$ ($0.96$, $0.95$, $0.95$). This indicates the model's resilience in predicting students who are likely to withdraw. The model's overall accuracy increases from $0.82$ to $0.98$ while course duration is considered $5\%$ to $100\%$, highlighting the growing dependability of predictions with additional course data, especially for the Pass and Withdrawn categories where scores regularly approach or surpass ideal levels. The overall accuracy, along with both macro and weighted averages, exhibits a comparable upward trend, highlighting the model's stability and reliability in forecasting student results as data availability increases during the course. The results demonstrates that the proposed model is suitable for accurate early prediction of student performance in an online learning environment.

\subsubsection{ROC with different course length}
\begin{figure}[h!]
	\centering
	\subfigure[Course duration $5\%$]{\includegraphics[width=0.49\linewidth]{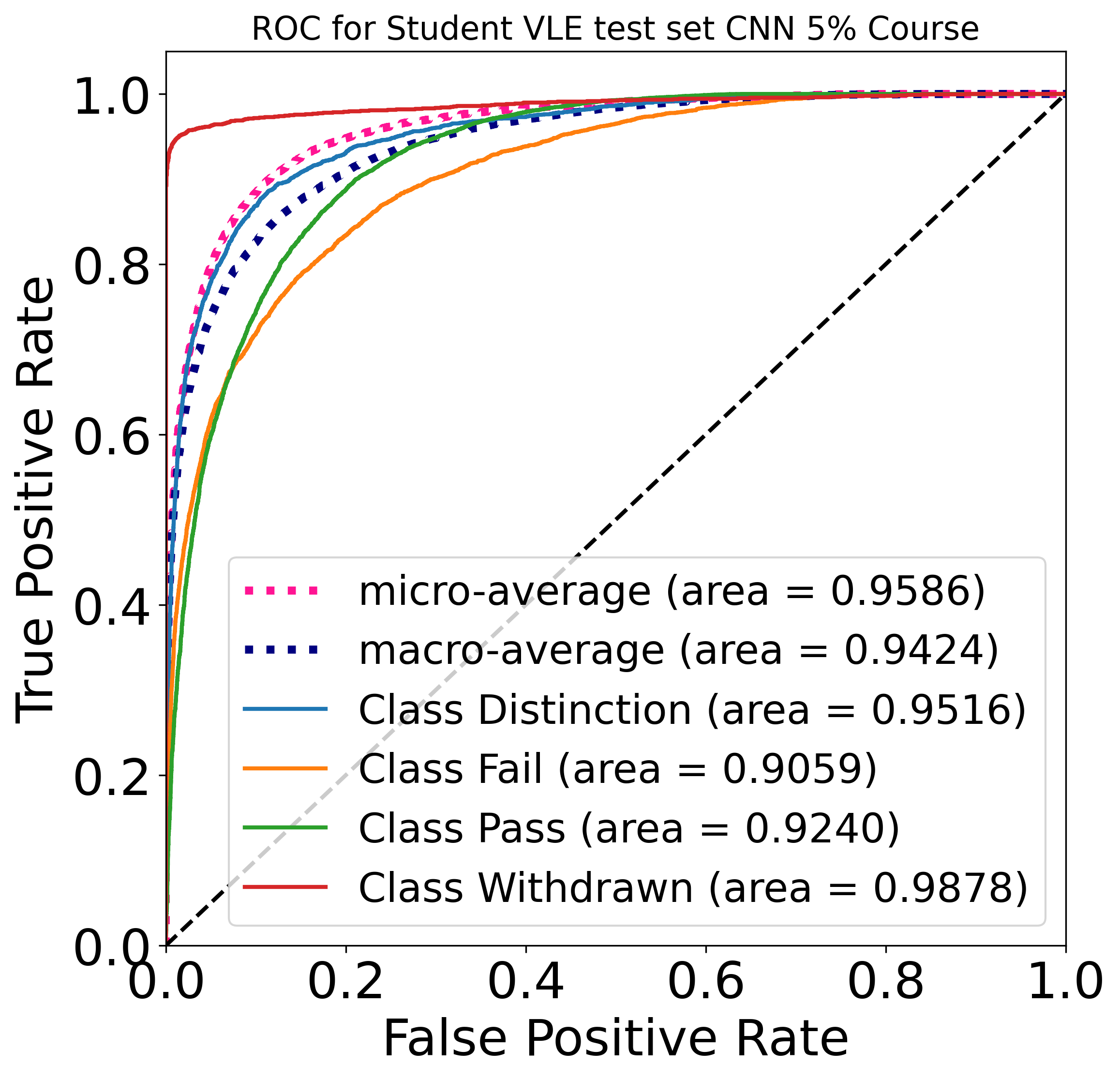}}
	\subfigure[$10\%$ of dataset]{\includegraphics[width=0.49\linewidth]{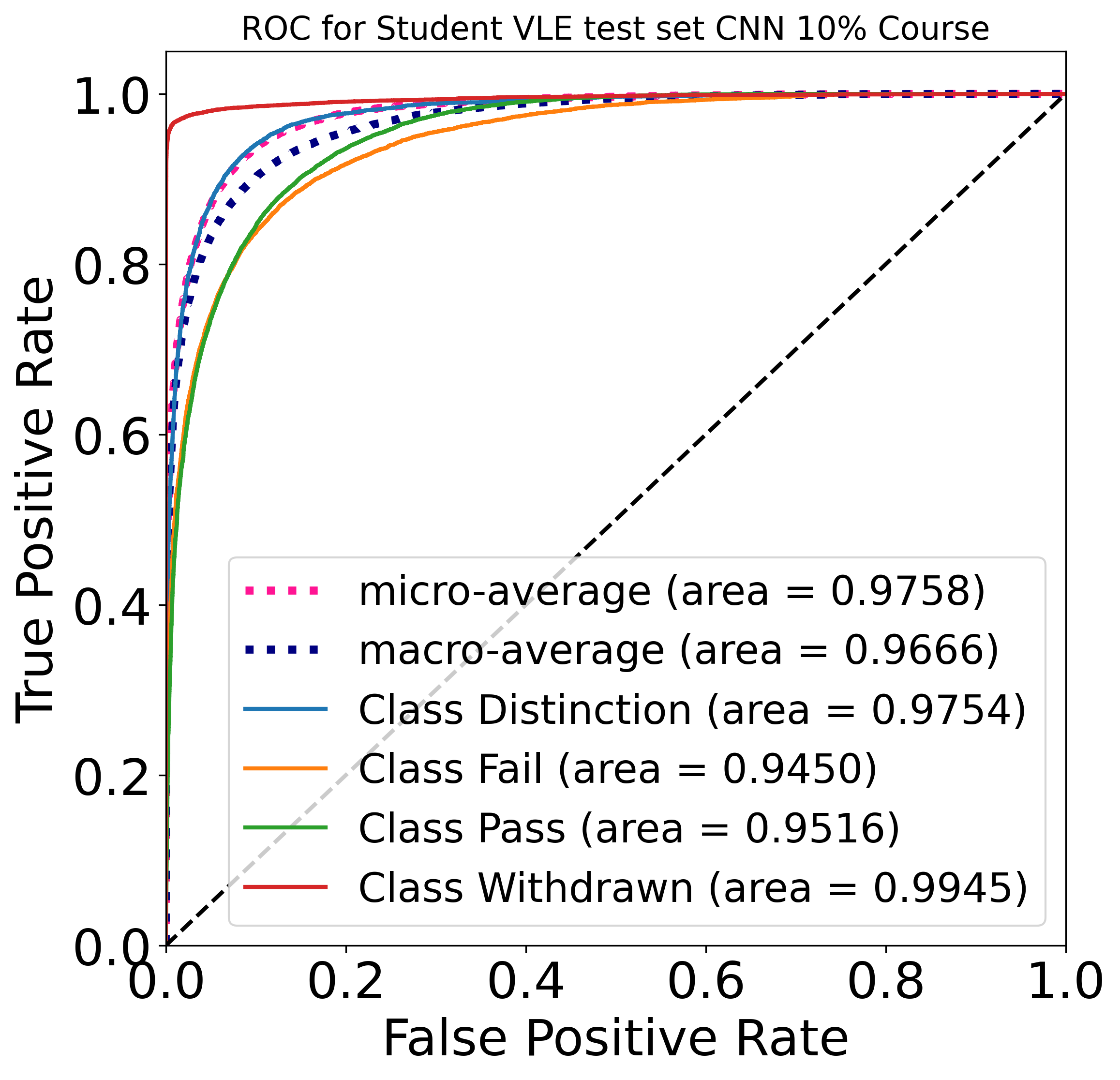}}
	\subfigure[Course duration $20\%$]{\includegraphics[width=0.49\linewidth]{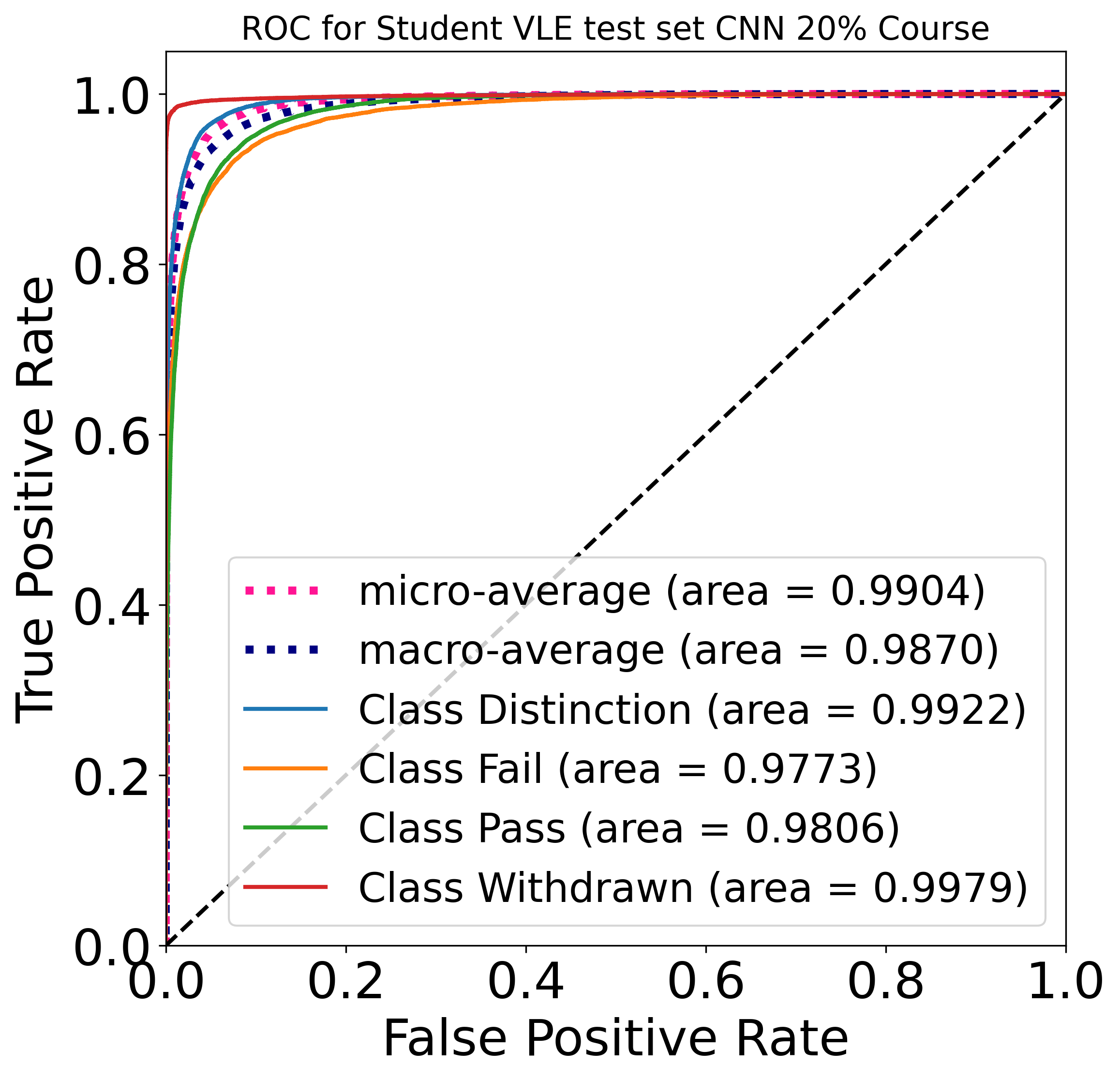}}
 	%\subfigure[$40\%$ of dataset]{\includegraphics[width=0.32\linewidth]{ROC for Student VLE test set CNN 40 Course_new.png}}
  	%\subfigure[$60\%$ of dataset]{\includegraphics[width=0.32\linewidth]{ROC for Student VLE test set CNN 60 Course_new.png}}
   	%\subfigure[Course duration $80\%$]{\includegraphics[width=0.32\linewidth]{ROC for Student VLE test set CNN 80 Course_new.png}}
    	\subfigure[Course duration $100\%$]{\includegraphics[width=0.49\linewidth]{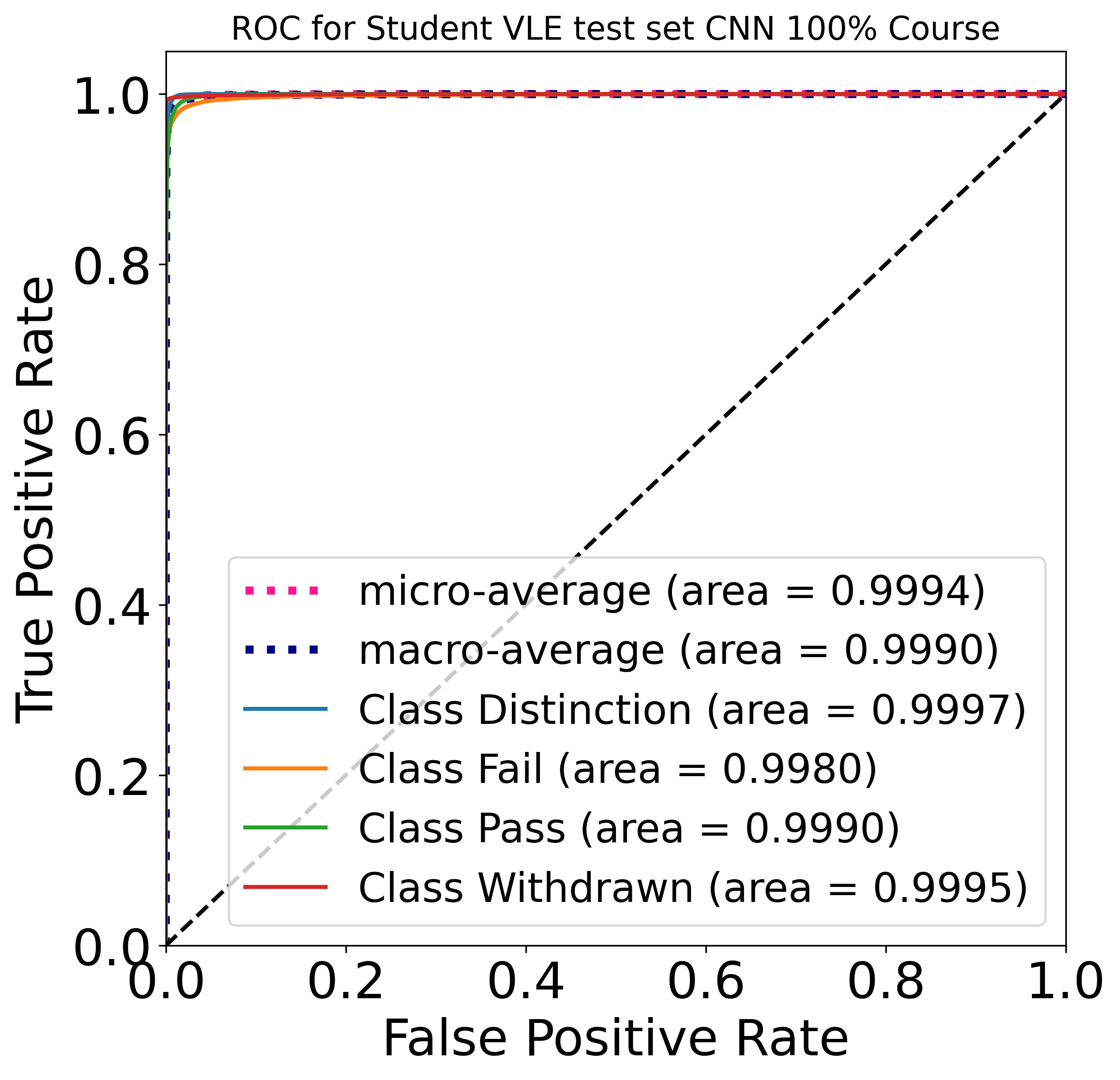}}
	\caption{Test set ROC for the proposed CNN at different course lengths: (a) $5\%$ (b) $10\%$ (c) $20\%$, and (f) $100\%$.}
	\label{fig:fig5 ROC curves}
\end{figure}

Figure \ref{fig:fig5 ROC curves} illustrates the ROC curve performance of a multiclass classification model across four classes: Distinction, Pass, Fail, and Withdrawn with different percentages of course duration ($5\%$, $10\%$, $20\%$, and $100\%$). Each solid line represents the ROC curve for one class, showing how well the model distinguishes that class from the others. The micro-average and macro-average curves are represented by dotted lines (pink and blue), which provide overall performance metrics, with AUC (Area Under the Curve) values of $0.9586$ and $0.9424$ while considering the $5\%$ of course duration, respectively, indicating strong overall classification ability as presented in Figure \ref{fig:fig5 ROC curves} (a). The individual AUC values for the classes range from $0.9059$ to $0.9878$, with the Withdrawn class showing near-perfect discrimination, highlighting the model's robustness in distinguishing between the different categories. As represented in Figure \ref{fig:fig5 ROC curves} (c), for the $20\%$ of course length, the AUC values for micro-average and macro-average curves are $0.9904$ and $0.9870$, respectively, which can be considered as a perfect early prediction of students' performance. Even the AUC curves values for Distinction, Fail, Pass, and Withdrawn are $0.9922$, $0.9773$, $0.9806$, and $0.9979$, respectively. As we consider the course length more than $20\%$, the model performance becomes accurate (almost $100\%$) in terms of individual AUC values for the classes as well as micro and macro-average as represented in Figure \ref{fig:fig5 ROC curves}. It is summarized from the results that the proposed model is very considerable for the accurate early at-risk prediction of students' performance in an online education environment, which will be very helpful for an intelligent education system. 

\subsubsection{Confusion Matrix with different course length}
\begin{figure}[h!]
	\centering
	\subfigure[Course duration $5\%$]{\includegraphics[width=0.49\linewidth]{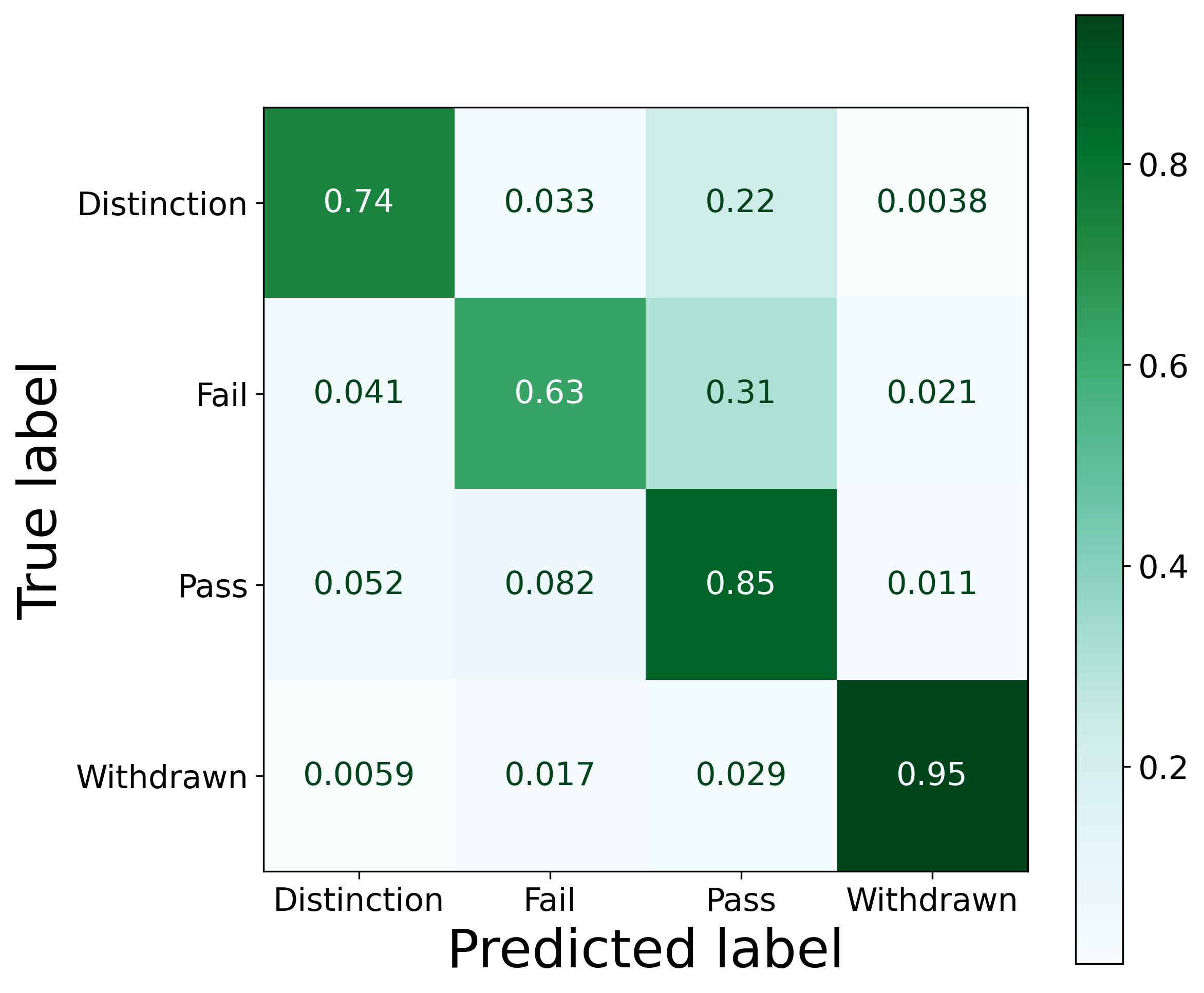}}
	\subfigure[$10\%$ of dataset]{\includegraphics[width=0.49\linewidth]{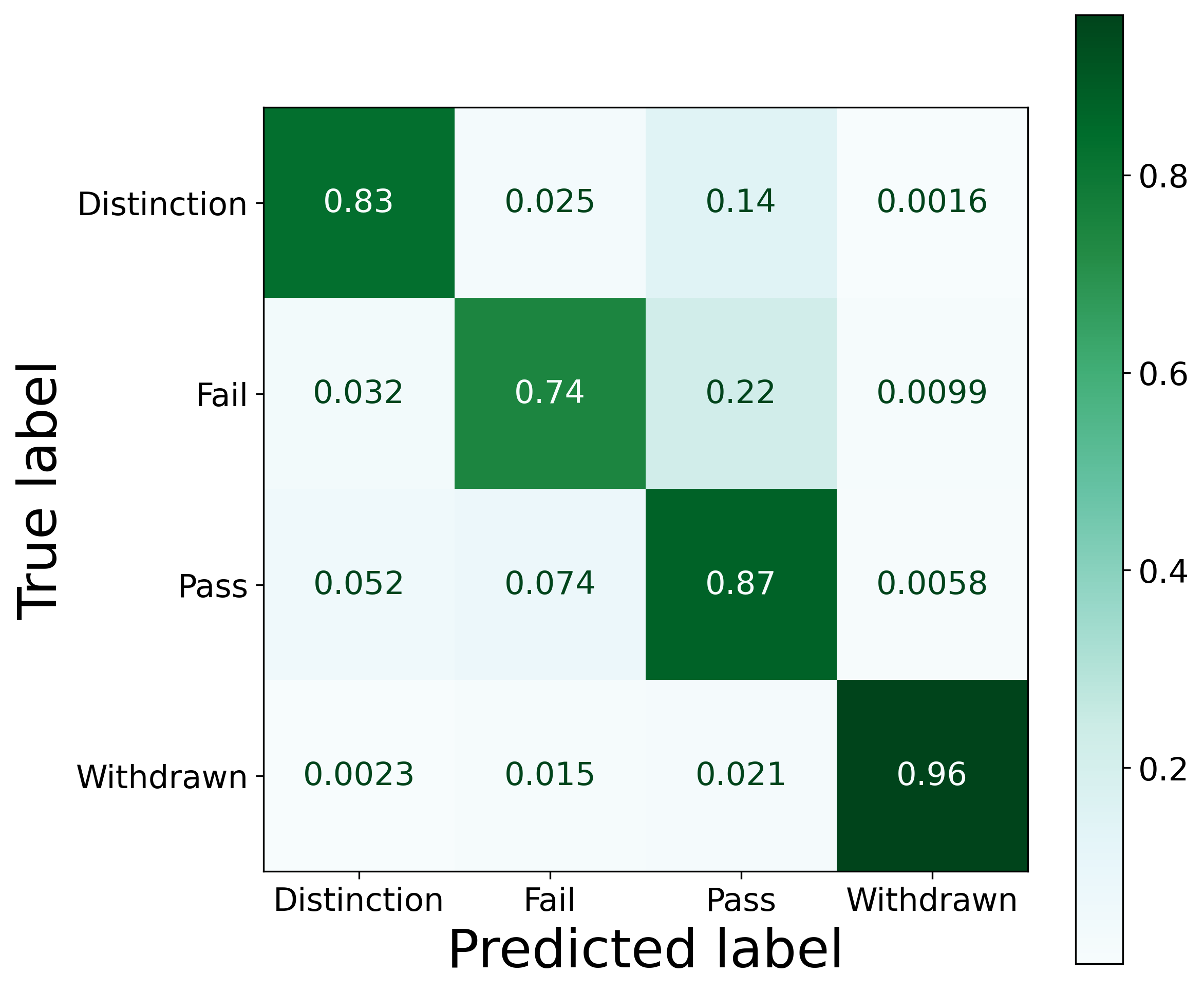}}
	\subfigure[Course duration $20\%$]{\includegraphics[width=0.49\linewidth]{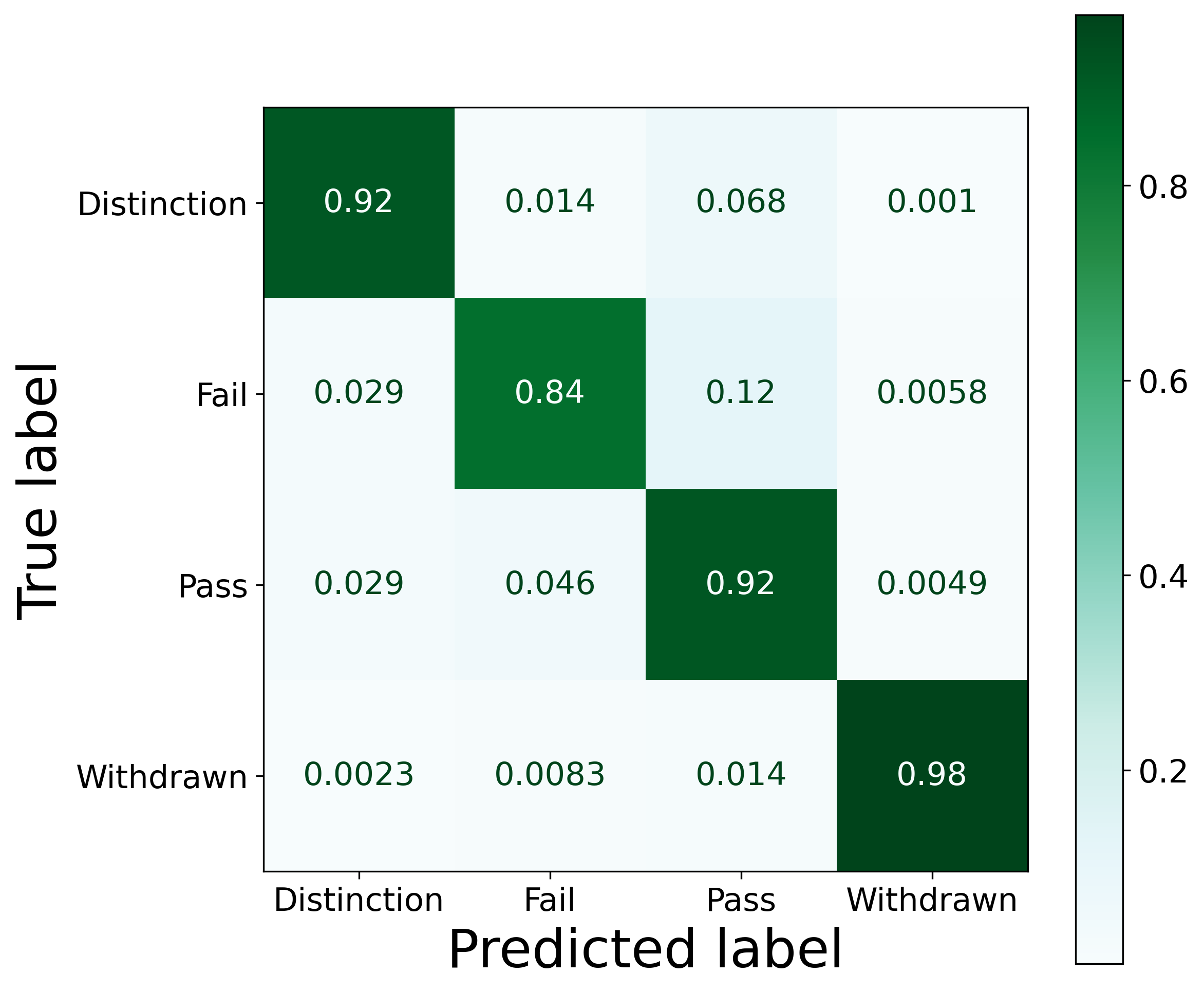}}
 	%\subfigure[Course duration $40\%$]{\includegraphics[width=0.32\linewidth]{Test set CNN 40 Course_new.png}}
  	%\subfigure[$60\%$ of dataset]{\includegraphics[width=0.32\linewidth]{Test set CNN 60 Course_new.png}}
   	%\subfigure[$80\%$ of dataset]{\includegraphics[width=0.32\linewidth]{Test set CNN 80 Course_new.png}}
    	\subfigure[Course duration $100\%$]{\includegraphics[width=0.49\linewidth]{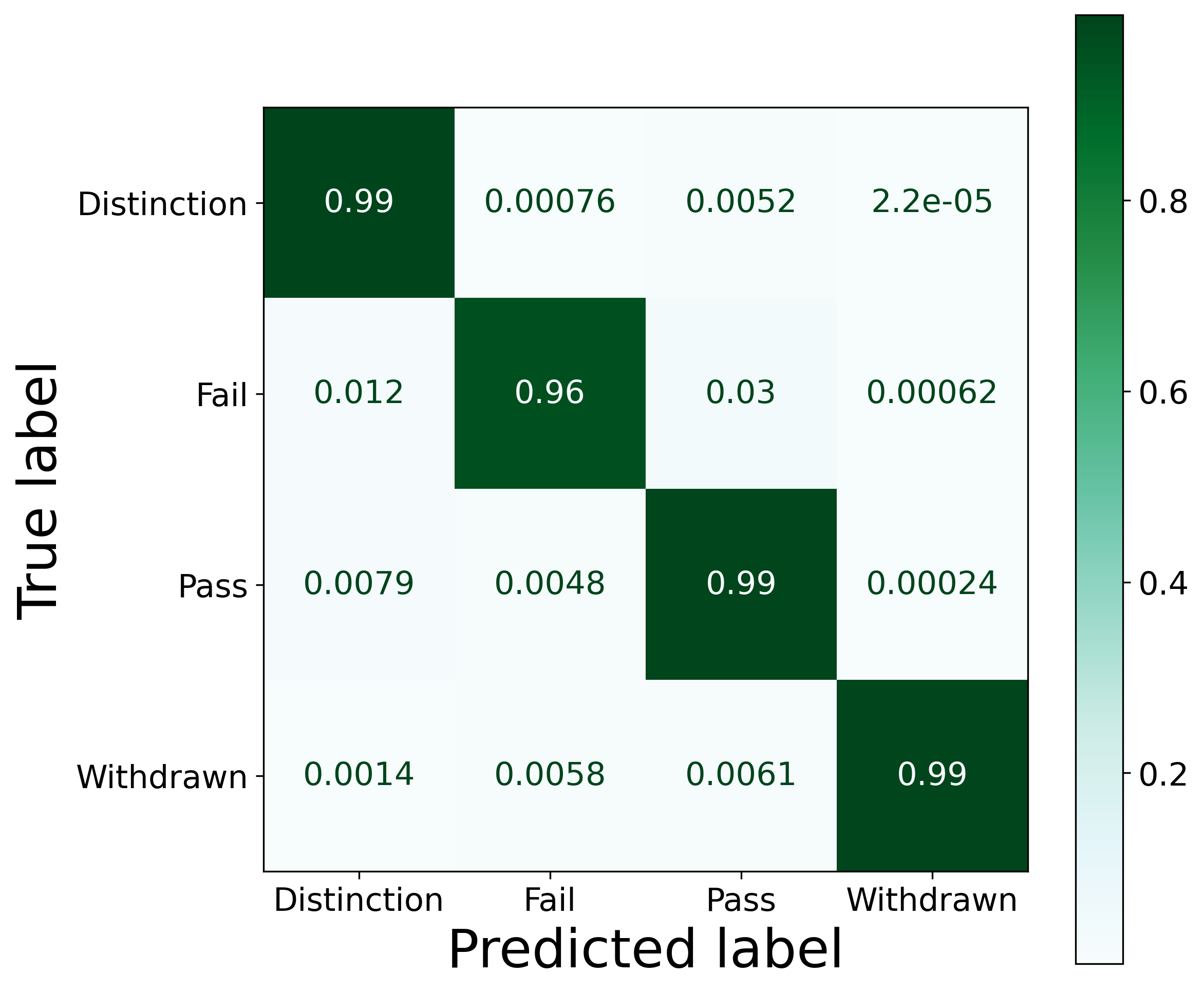}}
	\caption{Test set confusion matrices for the proposed CNN at different course lengths considering all features: (a) $5\%$ (b) $10\%$ (c) $20\%$ and (d) $100\%$.}
	\label{fig:fig5 confusion_matrix of all dataset}
\end{figure}

Figure \ref{fig:fig5 confusion_matrix of all dataset} presents confusion matrices with course duration of $5\%$, $10\%$, $20\%$, and $100\%$ while all features have been considered for training and testing process of the model. As the course length increases from $5\%$ to $10\%$, $20\%$, and $100\%$, the confusion matrix values in diagonals improve for all categories. The Withdrawn class value is exactly $0.96$, $0.98$, and $0.99$  while course duration is $10\%$, $20\%$, and $100\%$ as presented in Figure \ref{fig:fig5 confusion_matrix of all dataset} (b), (c), and (d). While the diagonal values of confusion matrices for other classes including Distinction, Fail, and Pass are almost the same at $20\%$ or more course lengths as depicted in Figure \ref{fig:fig5 confusion_matrix of all dataset} (c) and (d). The results suggested the proposed idea is suited to the virtual learning environment for the early prediction of students' performance.
\section{Conclusion}
In this study, we presented a novel approach to predicting early at-risk student performance in a Virtual Learning Environment (VLE) using a one-dimensional Convolutional Neural Network (1D-CNN) applied to the Open University Learning Analytics (OULAD) dataset. Our proposed model demonstrated superior performance compared to baseline models, including RF 'gini', RF 'entropy', DFFNN, and ANN-LSTM, across metrics such as accuracy, precision, recall, and F1-score.  Additionally, we evaluated the model's predictive accuracy at different stages of the course $5\%$, $10\%$, $20\%$, $40\%$, $60\%$, and $100\%$ revealing that at $20\%$ and $40\%$ of the course duration, the model achieves $92\%$ and $97\%$ accuracy, respectively. At the full course duration of $100\%$, with all features considered, the model attained an accuracy, precision, recall, and F1-score of $98\%$, underscoring its potential for accurately identifying at-risk students. These findings suggest that our methodology could significantly enhance early intervention strategies in online education platforms, enabling educators to proactively address students' needs.

%However, several limitations must be acknowledged. First, the model's success relies on the accuracy and completeness of the input data. Missing or erroneous data can have a substantial impact on the model's prediction power. Furthermore, while our model performed well on the OULAD, its applicability to different educational environments needs to be thoroughly evaluated. Although 1D-CNNs are well-suited for sequential data, they may constrain the model's capacity to capture more complex patterns and correlations, which could be better addressed by more sophisticated architectures.

For future research, we propose investigating the use of the Transformer-based models, which have recently shown higher performance in various domains such as natural language processing and time series analysis. Transformers, with their ability to capture long-term dependencies and complicated interactions in data, have the potential to improve student performance prediction by better understanding the intricate relationships between diverse features. Furthermore, a comparative analysis of the 1D-CNN model and the Transformer-based models could provide useful insights into each approach's strengths and limitations, enabling the development of more robust predictive models in education. This future research could result in even more accurate and generalizable models, ultimately leading to more personalized and effective educational interventions.

\section{Availability of Data and Material}
The dataset analysis for the experiments is publicly available on the website of the UK university:  
\href{https://analyse.kmi.open.ac.uk/open\_dataset}{https://analyse.kmi.open.ac.uk/open\_dataset}.

\section{Acknowledgement}
This work is supported by the Innovation Teams of Ordinary Universities in Guangdong Province (2021KCXTD038, 2023KCXTD022), Key Laboratory of Ordinary Universities in Guangdong Province (2022KSYS003), China University Indus-try, University, and Research Innovation Fund Project (2022XF058), Key Discipline Research Ability Improvement Project of Guangdong Province (2021ZDJS043, 2022ZDJS068), Hanshan Normal University Research platform project(PNB2104),
Hanshan Normal University Doctoral Startup Project (QD202127), and Hanshan Normal University Education and Teaching Innovation Project (HSJG-FZ21195).

\iffalse
\subsection{Acknowledgments}
We would like to thanks the School of Physics and Electronic Engineering, and Departemnt of Computer Science and Engineering, Hanshan Normal University, Chaozhou, China to provide the required resources and facilities. We would also like to thanks all authors for their important technical and writing suggestions.

\subsection{Conflict of Interest}
The authors declare that they have no conflict of interest.

\fi


\bibliography{manuscript}% common bib file

\end{document}